\documentclass{article} % For LaTeX2e
\usepackage{MLDD_workshop_2023,times}

\usepackage{amsmath,amsfonts,bm}

\def\eqref#1{equation~\ref{#1}}
\def\1{\bm{1}}

\DeclareMathAlphabet{\mathsfit}{\encodingdefault}{\sfdefault}{m}{sl}
\SetMathAlphabet{\mathsfit}{bold}{\encodingdefault}{\sfdefault}{bx}{n}

\usepackage{hyperref}
\usepackage{url}

\arxiv % Uncomment for arXiv deposit while under review.

\usepackage[utf8]{inputenc} % allow utf-8 input
\usepackage[T1]{fontenc}    % use 8-bit T1 fonts
\usepackage{booktabs}       % professional-quality tables
\usepackage{amsfonts}       % blackboard math symbols
\usepackage{nicefrac}       % compact symbols for 1/2, etc.
\usepackage{microtype}      % microtypography
\usepackage{xcolor}         % colors
\usepackage{amsmath}
\usepackage[bb=boondox]{mathalpha}
\usepackage{relsize}
\usepackage{graphicx}
\usepackage{bm}
\usepackage{algorithm}
\usepackage{algpseudocode}

\title{SupSiam: Non-contrastive Auxiliary Loss for Learning from Molecular Conformers}

\author{%
  Michael Maser\\
  Prescient Design, Genentech\\
  South San Francisco, CA \\
  \texttt{maserm@gene.com} \\
  \And
  Ji Won Park\\
  Prescient Design, Genentech\\
  South San Francisco, CA \\
  \texttt{parj2@gene.com} \\
  \And
  Joshua Yao-Yu Lin\\
  Prescient Design, Genentech\\
  South San Francisco, CA \\
  \texttt{liny82@gene.com} \\
  \And
  Jae Hyeon Lee\\
  Prescient Design, Genentech\\
  South San Francisco, CA \\
  \texttt{leej226@gene.com} \\
  \And
  Nathan C. Frey\\
  Prescient Design, Genentech\\
  South San Francisco, CA \\
  \texttt{freyn6@gene.com} \\
  \And
  Andrew Watkins\\
  Prescient Design, Genentech\\
  South San Francisco, CA \\
  \texttt{watkina6@gene.com} \\
}

\begin{document}

\maketitle

\begin{abstract}
  We investigate Siamese networks for learning related embeddings for augmented samples of molecular conformers. We find that a non-contrastive (positive-pair only) auxiliary task aids in supervised training of Euclidean neural networks (E3NNs) and increases manifold smoothness (MS) around point-cloud geometries. We demonstrate this property for multiple drug-activity prediction tasks while maintaining relevant performance metrics, and propose an extension of MS to probabilistic and regression settings. We provide an analysis of representation collapse, finding substantial effects of task-weighting, latent dimension, and regularization. We expect the presented protocol to aid in the development of reliable E3NNs from molecular conformers, even for small-data drug discovery programs.
\end{abstract}
%%%%%%%%%%%%%%%%%%%%%%%%%%%%%%%%%%%%%%%%%%%%%%%%%%%%%%%%%%
\section{Background \& Introduction} \label{sec:Intro}

Modeling conformational shape is of critical importance in many molecular machine learning (MolML) tasks \citep{Zheng}. This is especially true in the drug discovery (DD) regime, e.g., for predicting the affinity of a ligand-protein binding interaction \citep{Jones}. However, many programs in ML-based DD (MLDD) rely on small, noisy datasets ($O \, 10^{2-4}$) containing complex structures, making the development of generalizable 3D neural networks (NNs) particularly challenging.

Euclidean NNs (E3NNs) \citep{Geiger} make up the basis of many graph NN (GNN) models with equivariance to SE(3) transformations \citep{Liao, Batatia}. Atomic coordinates are used to define radial edges for spatial message passing, increasing GNN expressivity over covalent-only adjacencies \citep{Geiger}. Through-space interactions that intuitively influence structure activity relationships (SARs) \citep{Kombo, Sauer} are thus explicitly modeled. 

E3NNs have shown impressive performance for a variety of MolML tasks such as learning neural potentials \citep{Zaidi, Devereux} and predicting electronic properties \citep{Rackers, Thomas}. However, their use in drug-activity modeling is comparatively rare, likely owing to the data challenges described above. Furthermore, in this space, little is understood about the generalizability and latent properties of E3NNs. We seek to address this here and to better understand representation learning with molecular conformers.

\subsection{Motivation} \label{sec:Motivation}

Given the fundamental dependence of drug-target binding and SARs on 3D structure, we sought an understanding of E3NN behavior around molecular geometries. During supervised training with 3D datasets, we found that models were strikingly sensitive to input coordinates (see Section \ref{sec:pdc_ms}). This was seen as highly problematic, since models are often exposed to structures from varying experimental and/or computational methods in production settings. As such, we sought to develop learning methods to increase the generalizability of E3NNs to geometry perturbations. 

Inspired by recent works in self-supervised learning (SSL) for (Mol)ML \citep{Grill, Chen, Wang, Zaidi, Godwin}, we devised auxiliary tasks to aid in training smooth, supervised E3NN manifolds. Simple Siamese networks (SimSiam) \citep{Chen} were identified as a promising base method due to the following preliminary guidelines:

\begin{enumerate}
    \item We desire that networks embed very similar geometries (e.g., augmented pairs) to nearby latent vectors
    \item We do not require that distinct conformers of the same graph map to distant latent vectors (i.e., no negative pairs)
    \item Likewise, we do not require that conformations of distinct graphs (i.e., unique molecules) map to distant latent vectors 
\end{enumerate}

Guideline 1 is intuitively motivated; given the geometry dependence of biophysical interactions that create SARs, we desire models to learn that similar geometries should receive similar predictions. For 2 and 3, our decision to avoid negative pairing is based on two non-assumptions. First, we recognize that, in many cases, it \textit{will} be desirable for most conformers $c \in C$ to be embedded closely to one another. However, for many high-value tasks, we desire that networks precisely discriminate between individual conformers, as they may behave differently in biochemical systems. Second, molecules with unrelated connectivity (2D graph) can adopt very similar 3D shapes and possess similar \textit{functional} properties \citep{Sauer}. Therefore, though often reasonable, we avoid the assumption that is desirable for models to force embeddings of distinct molecules apart. 

Given this context, we developed positive-pair-only auxiliary tasks for 3D MolML. Presented herein are detailed investigations of \textit{Sup}ervised \textit{Siam}ese networks (\textit{SupSiam}) for MLDD tasks. 

%Comparison experiments that include negative-pair (CLR) tasks are included in Section \ref{sec:Results and Discussion} with discussion.

%%%%%%%%%%%%%%%%%%%%%%%%%%%%%%%%%%%%%%%%%%%%%%%%%%%%%%%%%%
\section{Related Work} \label{sec:Related}

\subsection{Multi-instance learning (MIL) with conformer ensembles (CEs)} \label{sec:MIL}

Despite the growing literature in equivariant GNNs, relatively little focus has gone toward the effects of 3D conformers themselves \citep{Axelroddqd, Ganearks, Ganea, Isert}. Existing efforts have even demonstrated a lack of performance gains from multiple-instance learning (MIL) with conformer ensembles (CEs) \citep{Axelroddqd}. Despite this, we expect that modeling the dynamics of CEs will be critical for many MLDD tasks, particularly for developing oracles that generalize to new 3D structures. To this end, we are unaware of detailed studies of MIL over CEs for activity-related tasks. Prior art in CE-MIL has focused only on electronic or quantum mechanical (QM) property prediction \citep{Axelroddqd, Zaidi, Godwin}, for which MolML has been well-demonstrated, even with 2D graphs \citep{Wu}.

\subsection{Contrastive learning (CLR)} \label{sec:CLR}

Contrastive learning (CLR) is in widespread use as a pre-training method for computer vision and other ML disciplines \citep{Le-Khac}. Recently, ``MolCLR'' was demonstrated to be effective for improving the performance of 2D GNNs in QM property prediction \citep{Wangynb}. Typical augmentation tasks include subgraph masking as well as node and/or edge dropout. Graphs augmented from the same parent molecule are treated as positive pairs, while those derived from different parents are treated as negative pairs. The pre-training objective is to minimize and maximize the embedding distance between positive and negative pairs, respectively. 

The rationale for this objective states that similar --- but different --- 2D graphs should map to similar latent vectors \citep{Le-Khac}. We pose that this treatment could be counterproductive given the fundamentals of SARs that we desire models to learn: Changing a molecule's connectivity (input) should change its properties (labels). This is especially concerning in the small-molecule setting (roughly $\leq$ 50 heavy atoms), where ``activity cliffs,'' or drastic label shifts arising from, e.g., single-atom edits, frequently plague DD programs \citep{Stumpfe, Aldeghi}. 

%Similarly, for the contrastive objective, it is possible to obtain identical or very similar augmented graphs from distinct examples in a dataset, particularly in MLDD where data points are often slight variations on a core scaffold. This presents the possibility that embeddings of very similar graphs are pushed apart, which is counter to the underlying principle of the positive-pair objective in MolCLR. We argue below that this is less problematic in the 3D regime, particularly for non-CLR.

\begin{figure}[t]
    \centering
    \includegraphics[width=0.98\textwidth]{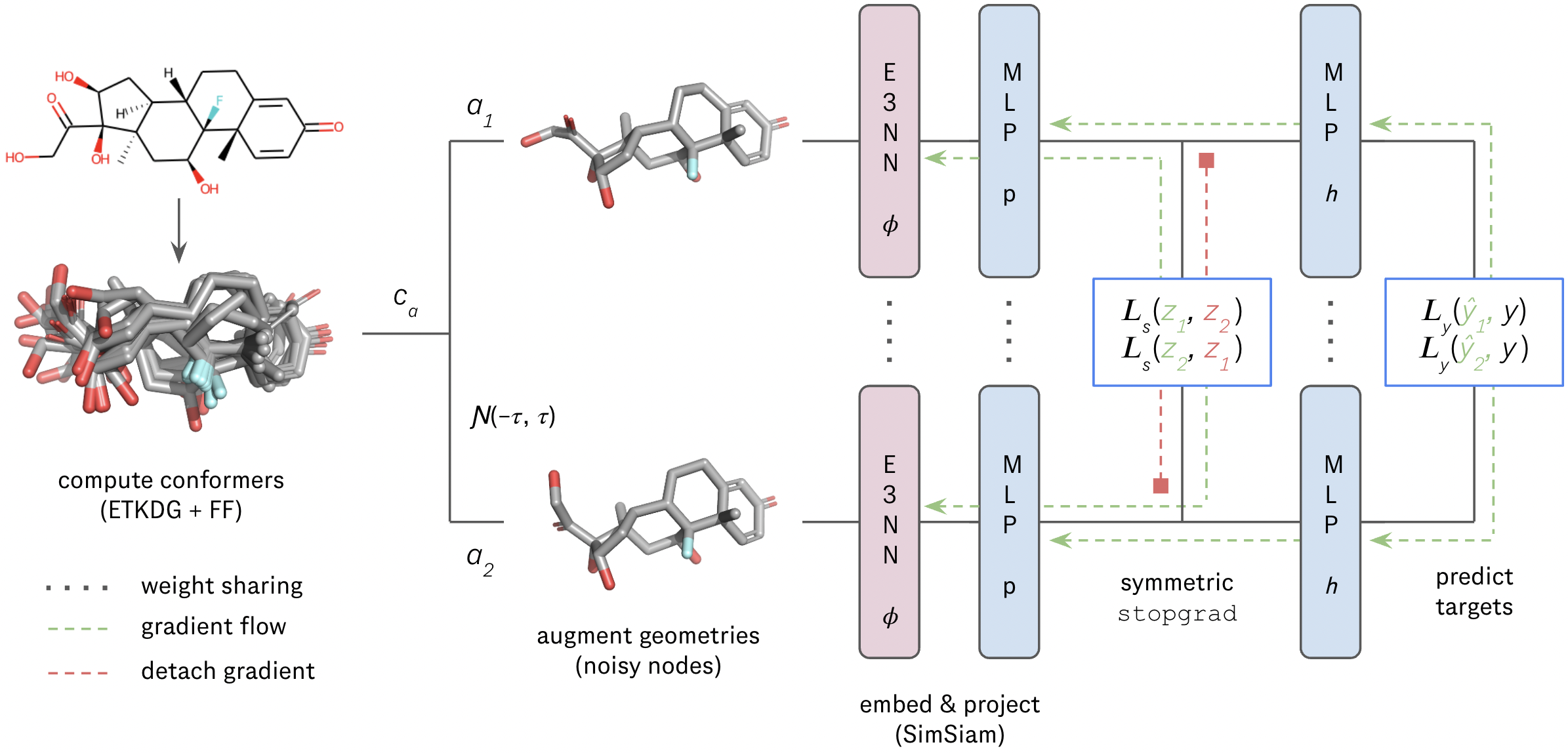}
    \caption{SupSiam pipeline.}
    \label{fig:pipeline}
\end{figure}

\subsection{Siamese networks (positive-only non-CLR)} \label{sec:SimSiam}

Non-CLR has been substantially developed for SSL/pre-training, in part to address the drawbacks of CLR above. SimSiam is an easily implemented non-CLR mechanism that is widely used in CV \citep{Chen}. As in CLR, augmented samples derived from parent inputs are treated as positive pairs, and models are trained to minimize the cosine distance between their embeddings. No negative pair constraints are imposed, satisfying our desired setting (see Section \ref{sec:Motivation}). 

A trivial global solution to the non-CLR task is to map all embeddings to a single latent point. To avoid such collapse, in SimSiam loss gradients are backpropagated only for one augmented sample and are detached from the rest. The loss is then symmetrized by multiple backward passes rotating the detached samples. The authors in \citet{Chen} claim that this \texttt{stopgrad} is sufficient to prevent collapse. They demonstrate that the embedding variance along the feature dimension remains roughly stable throughout training, indicating models are not learning to predict identical embeddings for all inputs. However, \citet{Li} recently showed that \textit{partial dimensional collapse} (PDC) can occur despite stable overall variance (see Section \ref{sec:Approach and Methods}).

\subsection{Pre-training by structure denoising}
Node denoising (``noisy nodes'') has recently been demonstrated as an effective pre-training task for modeling 3D graphs \citep{Godwin, Zaidi}. Input coordinates are augmented with Gaussian noise, and models are trained to predict this noise, i.e., to recover the ground-truth structure. 

While promising, this approach has two limiting pre-conditions: 1. Access to ground-truth (QM) conformations to denoise to and from; and 2. A suitably large corpus of QM structures of a relevant chemical space for pre-training. These conditions are quite difficult to satisfy for active DD programs, where structure optimization, e.g., by density functional theory (DFT) \citep{Kohn}, is prohibitively expensive, even for small labeled datasets. 

That said, including denoising as an auxiliary objective during supervised fine-tuning was shown to be beneficial, from which we take inspiration. At time of writing, we are unaware of the use of non-CLR tasks for CE-MolML, whether un-, self-, or label-supervised.

%
%%%%%%%%%%%%%%%%%%%%%%%%%%%%%%%%%%%%%%%%%%%%%%%%%%%%%%%%%%

\section{Approach and Methods} \label{sec:Approach and Methods}

A schematic of our approach is found in figure \ref{fig:pipeline}, and its components are described in detail in the following sections. All data and code are made available at \href{}{[link to be activated on acceptance for publication]}.

\subsection{Conformer preparation and augmentation}
Following \citet{Axelroddqd}, we prepare CEs of modest size ($C_m \leq 10$) for each molecule $m$ in a dataset with the inexpensive Experimental Torsion Knowledge Distance Geometry (ETKDG) method \citep{Wang}. 

We note that conformer selection for MolML remains an outstanding challenge \citep{Axelroddqd, Zaidi, Ganea}. In attempt to isolate the effects of the non-CLR mechanism from a dependence on starting conformers, we randomly sample a single conformer $c \in C_m$ for each molecule at each pass through models. This has the added benefit of computational efficiency over modeling a full CE in each pass. We report aggregated results over repeated training runs to marginalize over the effects of conformer selection. 

Following \citet{Godwin, Zaidi}, we augment conformers $c$ by sampling Gaussian noise $\mathcal{N}(0, 1) \in \mathbb{R}^{n \times 3}$ around normalized node positions $v_i^{c} \in V_c$ to give $c_a$. The noise scale is controlled by a temperature hyperparameter $\tau$ (i.e., noise is sampled from $\mathcal{N}(0, \tau)$). A cutoff radius of 4.0 \AA \, was used for constructing radial graphs, to which self-loops were added.
%Additional stochasticity can be added via noise dropout with probability $d$, though this is not studied here.

\subsection{Siamese E3NNs} \label{sec:siamese_e3nns}
E3NNs \citep{Thomas} were utilized as a base architecture to demonstrate our approach, though it is architecture-agnostic. The overall loss for optimization combines a target-prediction term ($\mathcal{L}_y$), a positive-only cosine embedding term ($\mathcal{L}_s$), and an $l_2$-regularization penalty ($\mathcal{L}_r$) as follows:

% = \lambda_u\mathcal{U}(y | x_{t}, z'_{t}) + \lambda_d\mathcal{D}(x_t, x_{t-1}) + \lambda_y \mathcal{L}_y(y_{t}, y)
\begin{equation} \label{eqn:loss}
    \mathcal{L} = \frac{1}{N} \sum_{i=1}^N \left ( \frac{1}{C_m} \sum_{c=1}^{C_m} \left [ \frac{1}{A} \sum_{a=1}^A \left ( \lambda_y \mathcal{L}_y(\hat{y}^c_{a}, y_i) + \lambda_r \mathcal{L}_r(z_{a}^c) \right ) + \frac{1}{A-1} \sum_{a=2}^A \lambda_s \mathcal{L}_s(z^c_1, z_{a}^c) \right ] \right ) \, ,
\end{equation}

\begin{equation} \label{eqn:cos}
    \mathcal{L}_s(z^c_1, z_{a\neq1}^c) = - \frac{1}{2} \left ( \left [ \frac{z^c_1}{\left\Vert z^c_1 \right\Vert_2} \cdot \frac{\xi(z_{a}^c)}{\left\Vert z_{a}^c \right\Vert_2} \right ] + \left [ \frac{z_{a}^c}{\left\Vert z_{a}^c \right\Vert_2} \cdot \frac{\xi(z^c_1)}{\left\Vert z^c_1 \right\Vert_2} \right ] \right ); \, \mathcal{L}_r(z_{a}^c) = \left\Vert  z_{a}^c \right\Vert_2 \, ,
\end{equation}

where $N$ is the dataset size, $C_m$ is the number of conformers of molecule $m$ modeled in each pass (1 herein), $\lambda_{y,r,s}$ are sub-task weights, $A$ is the number of augmented samples of each conformer (including parent, 2 herein), $z^c_1$ and $z_{a}^c$ are the learned embeddings for the parent and augmented conformers, respectively, $y_i$ and $\hat{y}_{a}^c$ are the ground-truth and predicted labels for molecule $i$ and augmented conformer $c_a$, respectively, and $\xi(\cdot)$ represents the \texttt{stopgrad} operation. 

We utilize a moderate-capacity, feed-forward E3NN, the trunk of which consists of 4 convolutional interaction blocks as defined in \citet{Geiger}. This is followed by global mean pooling over node features and a readout multi-layer perceptron (MLP) of length 1. \texttt{layernorm} is applied to each interaction block. The penultimate parent and augmented representations $\hat{z}^c_1$ and $\hat{z}_{a}^c$ are projected by another MLP $h$ to give $z^c_1$ and $z_{a}^c$ of dimension $d$. 

The Siamese task $\mathcal{L}_s$ in Equation \ref{eqn:cos} translates to predicting $z_{a}^c$ from $\hat{z}^c_1$ with $h$, and vice versa. In this mechanism, each backward pass only propagates through one sample ($z^c_a$ or $z^c_1$), with gradients detached from the other. This is symmetrized such that each augmentation $c_a \in A$ receives a backward pass.

For the target task $\mathcal{L}_y$ in Equation \ref{eqn:loss}, probabilistic inference was utilized to account for aleatoric uncertainty in the datasets \citep{Kendall}. Models thus output parameterized distributions over logits, from which we sample before appropriate activation and loss calculation.

Each experiment assesses E3NNs over 5 random weight instantiations, and experiments are run in duplicate (10 total runs). For test-set evaluation, weights of each ensemble member were loaded from the epoch of their best validation-set performance for the applicable metric (\textit{vide infra}).

\subsection{Datasets} \label{sec:datasets}

%%%%% WILL BE EXPANDED UPON %%%%%

It was hypothesized that our approach may be most useful for shape-based MLDD tasks such as binding/affinity prediction. For initial studies, one dataset both in and out of this task type were selected from the Therapeutic Data Commons (TDC) \citep{Huang}. These included one classification (\ref{sec:pgp}) and one regression (\ref{sec:clear}) dataset, for which ablation studies over $\tau$, $d$, and $\lambda_{s,r}$ were performed.

Dataset splits were used as given in the TDC API (\href{https://github.com/mims-harvard/TDC}{link}). For consistency, each dataset was restricted to the set of molecules containing only atoms of types \texttt{[C, N, O, F, S, Cl, Br, I]}. Compounds that failed conformer generation were also removed. Initial conformer diversity was imposed by RMSD $\geq$ 0.1 \AA. All conformers were optimized with the Universal Force Field (UFF) using default parameters in RDKit \citep{ny}.\footnote{It is possible that initial conformers converge to a small number of locally optimal geometries. We allow this under the assumption that this final set may be most reflective of that observed in a biological setting. I.e., since a random conformer is sampled in each epoch, models are roughly exposed to a Boltzmann-weighted distribution of conformations. Further study into \textit{learned} conformer sampling will be presented in future work.} 

Brief dataset details are given below, and we direct the reader to the TDC web page (\href{https://tdcommons.ai/single_pred_tasks/overview/}{link}) for full descriptions and original references.

\subsubsection{Pgp\_Brocatelli (\textbf{Pgp})} \label{sec:pgp}
This dataset comprises 1,212 molecules with affinity labels for binding to P-glycoprotein receptors. The task is binary classification; $\mathcal{L}_y$ is binary cross entropy (BCE), and the evaluated metric is ROCAUC.

\subsubsection{Clearance\_Hepatocyte\_AZ (\textbf{Clear})} \label{sec:clear}
This dataset contains 1,020 molecules with continuous labels for degree of hepatocyte clearance. The task is regression; $\mathcal{L}_y$ is mean squared error (MSE), and the evaluated metric is Spearman $\rho$.

\subsection{Analysis and metrics} \label{sec:analysis}

We hypothesized that the auxiliary task in \ref{eqn:loss} could result in more generalizable E3NNs in small-data regimes. This was analyzed by quantification of \textit{local manifold smoothness} (MS, $\eta_f$) \citep{Ng} as a proxy for model $f$'s robustness to conformer noise in unseen data (see Section \ref{sec:Motivation}). 

In \citet{Ng}, $\eta(f, c)$ is defined as the total percentage of augmented samples $c_a$ of input $c$ that are assigned the mode predicted label in the set (for binary tasks). We generalize this to the probabilistic and regression settings by computing the KL divergence between predicted posterior distributions $(\hat{\mu}^{c}_1, \hat{\sigma}^{c}_1)$ and $(\hat{\mu}_{a}^{c}, \hat{\sigma}_{a}^{c})$ for parent and augmented samples, respectively:

\begin{equation} \label{eqn:MS}
    \eta_f = \frac{1}{N} \sum_{n=1}^N \frac{1}{C_m(A-1)} \sum_{i=1}^{C_m} \sum_{a=2}^A 1 - \left [ \log(\hat{\sigma}_{a}^{c}) - \log(\hat{\sigma}^{c}_1) + \frac{(\hat{\sigma}^{c}_1)^2 + \left (\hat{\mu}_{a}^{c} - \hat{\mu}^{c}_1 \right )^2}{2(\hat{\sigma}_{a}^{c})^2} - 0.5 \right ] .
\end{equation}

While difficult to assess on absolute scale, we utilize $\eta_f$ to compare between models with different weightings of the subtasks and noise hyperparameters. 
%Since in some TDC datasets the default splits are such that test samples are roughly i.i.d. with training samples, we additionally evaluate $\eta_f$ in another holdout set comprising FDA approved drugs with 3D structures from the e-Drug3D database \citep{Douguet}.

\begin{figure}[t]
    \centering
    \includegraphics[width=0.98\textwidth]{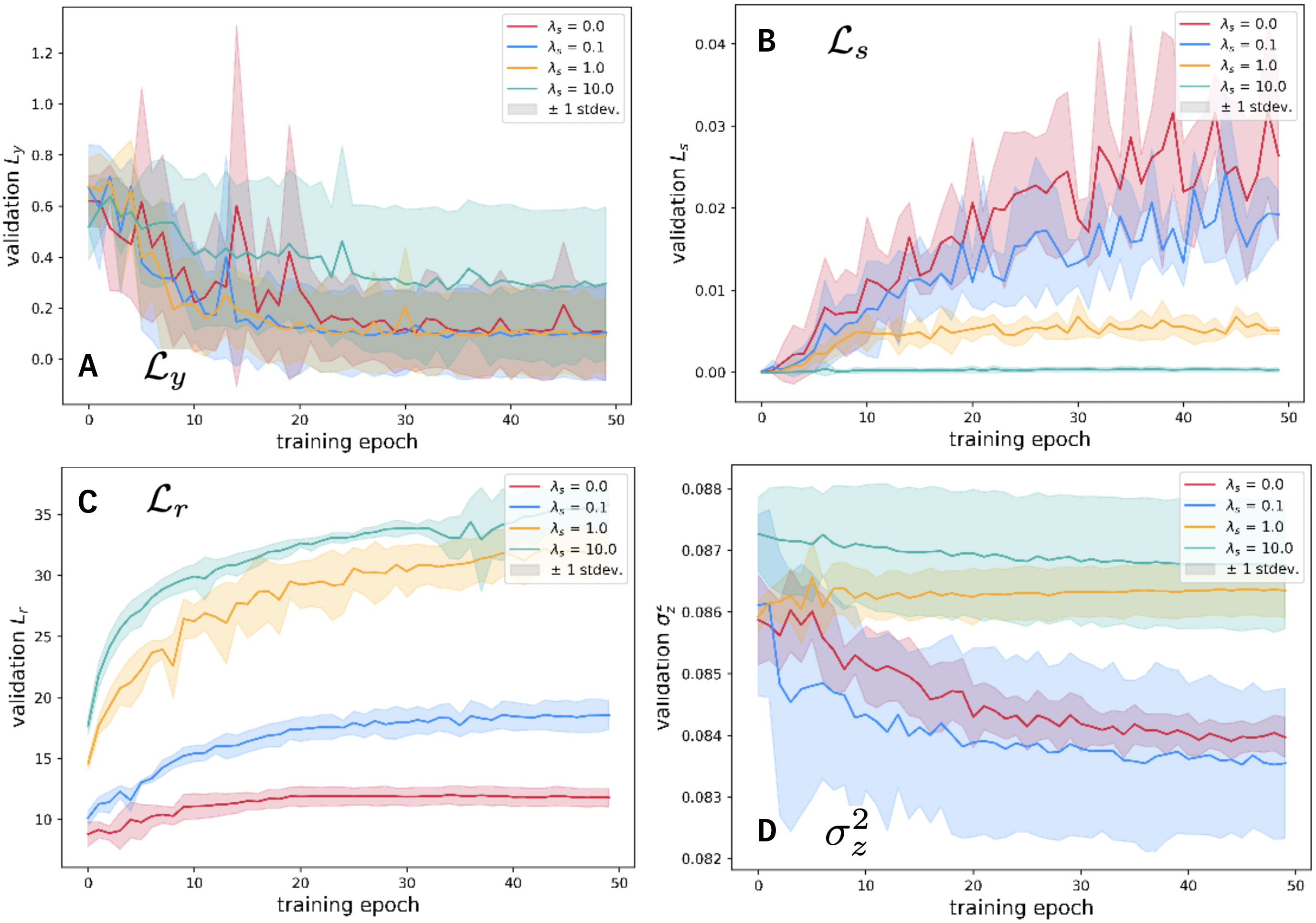}
    \caption{E3NN Pgp-binding classification training profiles at varying $\lambda_s$. A) target loss (BCE); B) Siamese loss (cos); C) regularization loss ($l_2$-norm); D) embedding feature variance. All curves show validation set results with $A = 1$, $d = 128$, $\tau = 0.1$, $\lambda_r = 0.0$.}
    \label{fig:curves}
\end{figure}
\begin{figure}[t]
    \centering
    \includegraphics[width=0.98\textwidth]{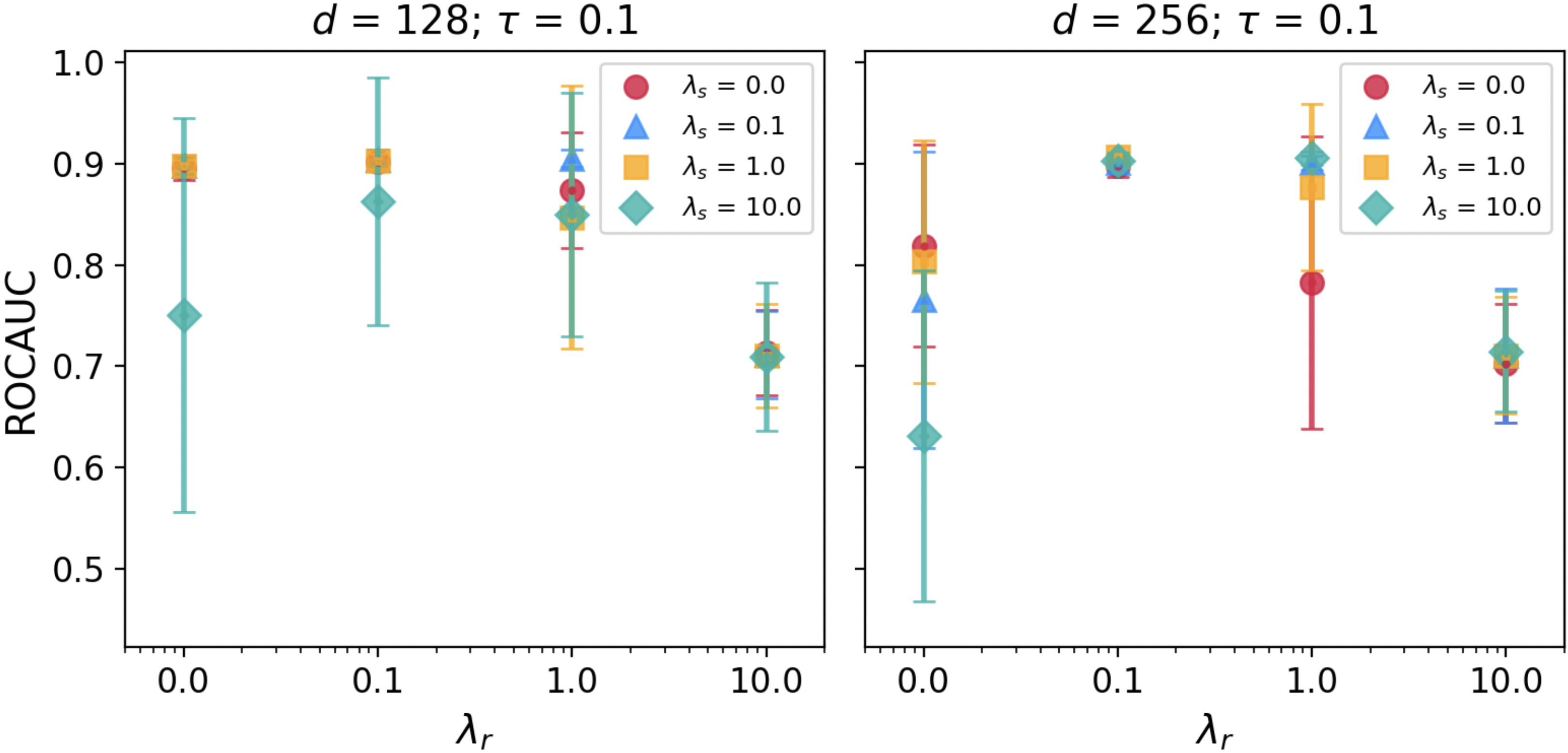}
    \caption{E3NN Pgp-binding classification test set ROCAUC across SupSiam parameters.}
    \label{fig:rocauc}
\end{figure}

%For fairness given the possibly substantial domain shift, this is only assessed for sake of comparison between models with varying $\lambda_{y,s,r}$.

Following \citet{Chen}, we detect trivial collapse by quantifying the variance in embeddings ($\sigma^2_{z}$) along the feature axis (see section \ref{sec:SimSiam}). Finally, we follow \citet{Li} to detect \textit{partial dimensional collapse}. We quantify the cumulative explained variance (CEV, $\Gamma$) of the singular values $\gamma$ computed via principal component analysis of embedding sets. The CEV up to rank-sorted $\gamma_j$ ($\Gamma_j$) and the area under the full CEV curve ($\bm{\Gamma}$) are defined as:

\begin{equation} \label{eqn:CEV}
    \Gamma_j = \frac{\sum_{i=1}^{j} \gamma_i}{\sum_{k=1}^{d} \gamma_k} \, ; \, \, \bm{\Gamma} = \frac{1}{d} \sum_{j=1}^{d} \Gamma_j \,,
\end{equation}

where $d$ is the full embedding size. $\bm{\Gamma}$ ranges between [0.5, 1.0]; larger values correspond to more rapid coverage of the total CEV over fewer singular values, and thus indicate a larger degree of collapse. $\bm{\Gamma} = 0.5$ corresponds to zero PDC.

%For each experiment, metrics and training profiles were evaluated via grid search over $\lambda_{y,s,r}$ and $\tau$ values. Given recent findings on the effect of model capacity on Siamese networks \citep{Li}, a range of hidden dimensions $d$ was additionally tested (see Supplementary for detailed results).

%\begin{figure}[t]
%    \centering
%    \includegraphics[width=0.99\textwidth]{SupSiam/Pgp_Broccatelli_roc_auc_score_all.png}
%    \caption{E3NN Pgp-inhibition classification ROCAUC scores vs $\lambda_s$ \& $\lambda_r$; aggregated over 15 initializations at varying $d$ \& $\tau$}
%    \label{fig:scatter}
%\end{figure}

\section{Results and Discussion} \label{sec:results}

\subsection{Training profiles} \label{sec:training}

Figure \ref{fig:curves} shows E3NN training profiles on the \textbf{Pgp} task at varying $\lambda_s$ values without regularization ($\lambda_r = 0.0$). Target loss curves are largely consistent across $\lambda_s$ values, (Figure \ref{fig:curves}A), though this is not the case in all settings (see Section \ref{app:sec:training} for full results). It is worth noting that with standard supervision ($\lambda_s = \lambda_r = 0$), training curves are often highly erratic. Further, cosine embedding distance for augmented pairs actually \textit{diverges} over the course of training for purely supervised models (Figure \ref{fig:curves}B, red). As anticipated, this divergence is mitigated using SupSiam, with an intuitive trend at increasing $\lambda_s$. At higher levels of $\lambda_s$, smooth training toward $\mathcal{L}_s = 0$ is observed.

Interestingly, the opposite trend is observed for $\mathcal{L}_r$ (Figure \ref{fig:curves}C). However, this only holds without regularization ($\lambda_r = 0$); with $\lambda_r > 0$, $\mathcal{L}_r$ smoothly converges in all cases (see Section \ref{app:sec:training}).

%At the same time, latent feature variance decreases monotonically, though this is not true in all experiments and in many cases is to a lesser degree than with $\lambda_s > 0$ (Figure \ref{fig:curves}D). Despite all this, the benchmark metric (ROC AUC score) converges (Figure \ref{fig:curves}D), although slowly, and in some cases converges to state of the art (SOTA) performance levels (see Supplementary). We find this type of modeling profile alarming and expect severe over- and/or mis-fitting to undesired data features (see Section \ref{sec:pdc_ms} for more).

%These training behaviors are markedly different with inclusion of the auxiliary task. Smoother loss curves are seen under many (but, importantly, not all) hyperparameter settings (Figure \ref{fig:curves}A).  
 Surprisingly, embedding-feature variance decreases monotonically over training at low values of $\lambda_s$, and is largely flat at higher $\lambda_s$ (Figure \ref{fig:curves}D). This effect is dependent on regularization. With $\lambda_r = 0$ as in Figure \ref{fig:curves}, the maximum $\lambda_s = 10$ actually maintains the highest feature variance throughout training. With $\lambda_r > 0$, however, $\sigma_z^2$ behaves as observed in \citet{Chen} --- $\sigma_z^2$ sharply decreases in early epochs before recovering to inital levels and plateauing (see Section \ref{app:sec:training}).
 
 %Finally, convergence of the benchmark metric can be maintained at lower $\lambda_s$, but does deteriorate over a certain (undetermined) threshold (Figure \ref{fig:curves}D). Under many settings, this again reaches SOTA levels and even outperforms the pure supervision setting ($\lambda_s = 0$). 

%Interestingly, increasing $\lambda_r$ past a critical point has a uniformly deleterious effect on training and performance (see Supplementary). In many cases at $\lambda_r \geq 10$, loss curves are completely flat and performance metrics hover around baseline (random) throughout fitting. 

\subsection{Performance metrics} \label{sec:performance}
Final test set performances are shown in Figure \ref{fig:rocauc}, with error bars representing $\pm$ 1 stdev. over repeat runs. In line with the observations above, test set ROCAUC is highest at intermediate levels of $\lambda_s$ and $\lambda_r$, which holds across settings of $d$. At small $\tau$, reasonable performance can be maintained even at maximum $\lambda_s$. At maximum $\lambda_r$, however, the target task is only minimally learned, regardless of all other settings (see Section \ref{app:sec:training}).

It is important to note that for this task, the maximum ROCAUC obtained ($\sim 0.91$) is slightly below the literature benchmark ($\sim$ 0.95) \citep{Huang}. However, benchmark methods largely utilize tree-based learning with cheminformatic representations; modeling this dataset with E3NNs could be expected to be quite challenging. This is much in line with our motivation (see Sections \ref{sec:Intro}, \ref{sec:Motivation}), and exceeding benchmark metrics is neither a goal nor expectation. Altogether, we find the combination of performance with stable, physically reasonable training profiles obtained with SupSiam (Section \ref{sec:training}) to be compelling for use in production settings. 

The ablation of $\lambda_r$ and $\lambda_s$ provides insight into their effects on latent properties. We expect that $\mathcal{L}_{s}$ is not simply serving to compactify latent space (like $\mathcal{L}_{r}$) due to the following observations:

%This is not true
%There are occasional similarities in the effects of $\lambda_r$ and $\lambda_s$ on $\mathcal{L}_{s}$ and $\sigma^2_{z}$. Like with $\lambda_{s}$ (above), $\mathcal{L}_s$ more rapidly converges at increasing $\lambda_{r}$, even with $\lambda_{s} = 0$. This could be interpreted such that the $\mathcal{L}_{s}$ task is, to some extent, serving simply to compactify the latent space (much like $\mathcal{L}_{r}$) without actually learning information about similar conformers. Though possible (and even reasonable \citep{Li}), we expect this is unlikely the entire function of $\mathcal{L}_{s}$ due to the following observations:

\begin{enumerate}
    \item Maximizing $\lambda_r$ inhibits training of $\mathcal{L}_{y}$, regardless of $\lambda_s$ (Figure \ref{fig:rocauc}). Conversely, maximizing $\lambda_s$ results in viable training profiles at several settings of $\lambda_r$ (Figures \ref{fig:curves}A, \ref{fig:rocauc}, Section \ref{app:sec:training});
    \item Ablating $\lambda_s$ is uniformly deleterious to $\mathcal{L}_{s}$ across $\lambda_r$ values. Conversely, $\lambda_r$ has little to no effect on $\mathcal{L}_{s}$, regardless of $\lambda_s$ (Figure \ref{fig:curves}B, Section \ref{app:sec:training});
    \item As expected, at fixed $\lambda_s$, $\mathcal{L}_{r}$ decreases with increasing $\lambda_r$ (Section \ref{app:sec:training}). Conversely, at fixed $\lambda_r$, $\mathcal{L}_{r}$ actually tends to increase with increasing $\lambda_s$ (Figure \ref{fig:curves}C).
\end{enumerate}

These observations point to differing behaviors of Siamese learning and latent regularization. It appears that while $\mathcal{L}_{r}$ does compactify latent space (lower $l_2$), this does not necessarily push \textit{related} embeddings to closer cosine distances. Conversely, while $\mathcal{L}_{s}$ does push related embeddings to closer cosine distances, it does so with an increased expansion of latent space (higher $l_2$). We thus expect a task-specific balance of $\lambda_s$ and $\lambda_r$ may lead to the most desirable model properties.

% Comment on saturated benchmarks vs legit generalizability?

\subsection{Manifold smoothness and partial dimensional collapse} \label{sec:pdc_ms}

%To test whether the training profiles in Section \ref{sec:training} lead to greater E3NN generalization, MS and PDC were evaluated for all experiments (see Section \ref{sec:analysis}). 

\begin{figure}[t]
    \centering
    \includegraphics[width=0.95\textwidth]{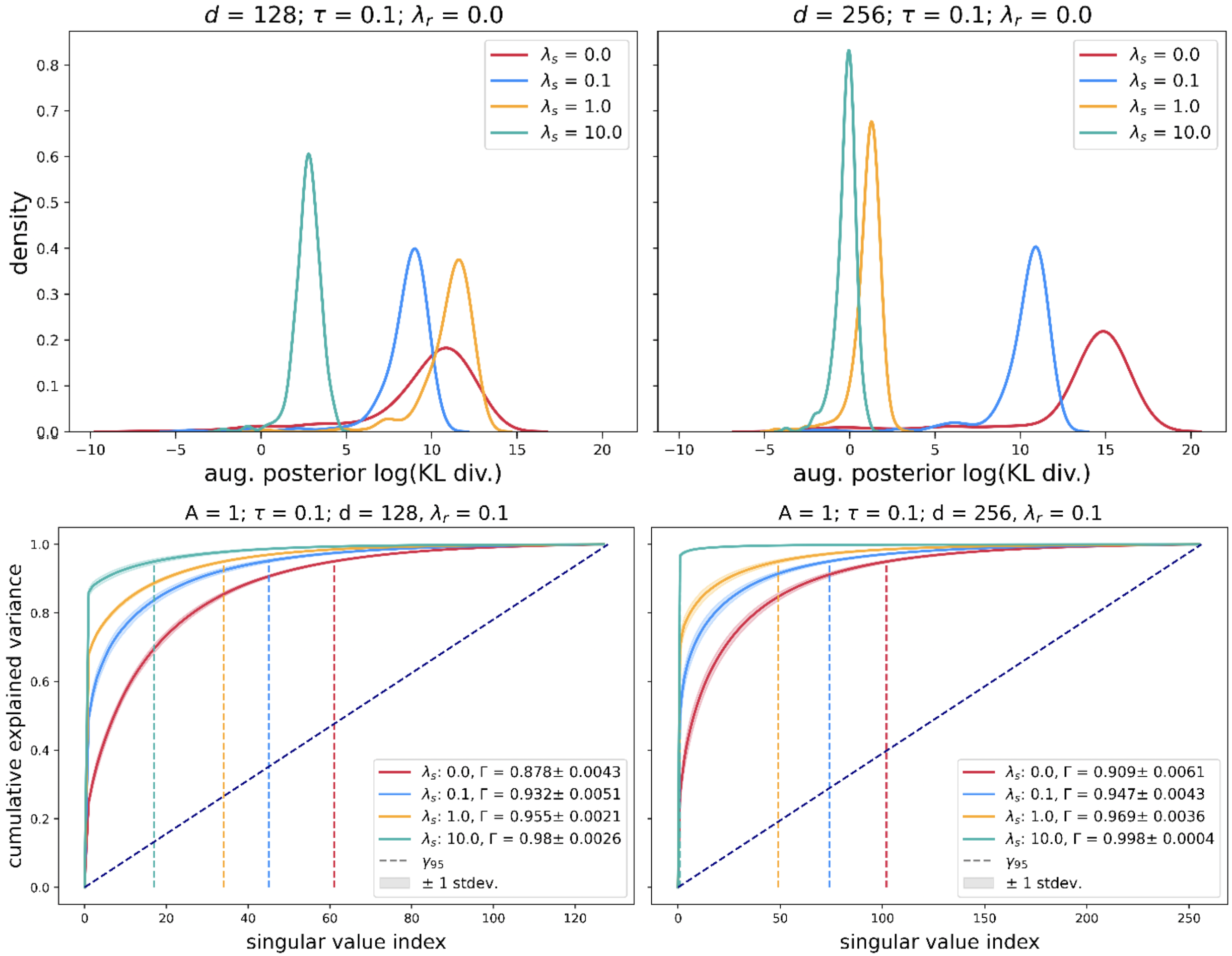}

    \caption{(Top) Pgp E3NN manifold smoothness ($\eta_f$) at varying $\lambda_s$. All models trained with $\tau = 0.1$, $A = 1$. $\eta_f$ evaluated with $\tau = 0.1$, $A = 10$. (Bottom) Pgp E3NN partial dimensional collapse at varying $\lambda_s$. Vertical lines indicate singular value at which $\Gamma_j = 0.95$ ($\gamma_j^{95}$).}
    \label{fig:ms_pdc}
\end{figure}

Figure \ref{fig:ms_pdc} (top) shows KDEs of per-molecule MS (Equation \ref{eqn:MS}) across $\lambda_s$ for the models discussed above. In many cases, SupSiam caused drastic reduction in posterior KL divergence between augmented samples, often by several log units. We note that the correlation of $\mu_f$ and $\lambda_s$ is often obscured when $\mathcal{L}_r$ is heavily weighted (see Section \ref{app:sec:ms_pdc}). In any case, we find MS analysis very informative. 

%It is important to note that the absolute scale of KL divergence reduces drastically at increasing $\lambda_r$ (see Supplementary). This intuitively reflects latent space compactification at high $\lambda_r$. That said, the trends across $\lambda_s$ remain largely unchanged.

 %\begin{figure}[t]  
 %\centering
 %\includegraphics[width=0.95\textwidth]{SupSiam/pdc.png}
%    \caption{Pgp E3NN partial dimensional collapse at varying $\lambda_s$. Vertical lines indicate singular value at which $\Gamma_j = 0.95$ ($\gamma_j^{95}$).}
%    \label{fig:pdc}
%\end{figure}

Figure \ref{fig:ms_pdc} (bottom) shows CEV curves across $\lambda_s$. We do observe a positive (albeit intuitive) correlation between $\bm{\Gamma}$ and $\lambda_s$. That said, in many cases only minor increases in PDC were observed at low levels of $\lambda_s$ vs $\lambda_s = 0.0$. It is striking to note the degree of PDC in the pure supervision setting ($\lambda_s = 0$). This may indicate that supervised E3NNs fit to relatively few input structure features, at least in small-data regimes. We find this alarming, but note that in most models (at $\lambda_s \leq 1$) 95\% CEV ($\gamma_j^{95}$) is not reached until roughly the $1/3$ or $1/2$ embedding index (Figure \ref{fig:ms_pdc}, bottom).

Contrary to \citet{Li}, we also observe a positive correlation between $\bm{\Gamma}$ and $d$. In \citet{Li}, increasing model capacity (via $d$) was used explicitly to \textit{reduce} PDC. This was rationalized as insufficient capacity resulting in information loss. It is possible that, at least in our data setting ($N \sim 10^{2-4}$), the opposite may be true, where an encoded-information limit is reached at relatively small $d$, and thus increasing capacity actually results in higher $\bm{\Gamma}$s. Increasing $\lambda_r$ decreased the dependence on $\lambda_s$, and in some cases actually reduced PDC (Section \ref{app:sec:ms_pdc}).

\citet{Li} propose continual learning as a mechanism to reduce PDC in Siamese networks. We leave such studies in our setting for future works. For now, we assess that low $\lambda_s$ settings lead to acceptable $\Delta \bm{\Gamma}$ from pure supervision, where collapse appears to be an outstanding issue. We expect that PDC may be a useful tool to analyze learned information in MolML more broadly.

\section{Conclusion and Outlook}

\textit{Sup}ervised \textit{Siam}ese networks (\textit{SupSiam}) were presented as an approach to train E3NN models on molecular geometries. We observe that the Siamese auxiliary task results in desirable latent properties while maintaining good performance in target prediction. Additionally, in many cases SupSiam was shown to increase manifold smoothness (i.e., decrease model sensitivity) to noise in input structures, a critical challenge in MolML. Lastly, embedding collapse was observed to increase only slightly over pure supervision. That said, the partial dimensional collapse of E3NNs was observed to be severe in most cases, which we find alarming. Future works will seek to tackle this issue.

% Not true anymore
%Critically, it is observed that training of the target task ($\mathcal{L}_{y}$) is unstable under \textit{all} hyperparameter settings without the Siamese task (i.e., $\lambda_s = 0$). To be clear, SupSiam does not ameliorate this behavior under all settings of $\lambda_s > 0$. However, stable training \textit{can} be achieved under auxiliary supervision, and we have yet to observe this without it. Altogether, the modeling behaviors using SupSiam reflect both chemical and ML intuition, which we find desirable and more trustworthy. 

Other topics for future work include expansion to other equivariant architectures \citep{Batatia} and augmentation mechanisms, e.g., noising bond distances and angles instead of atomic positions. Further, we are actively exploring non-CLR pre-training mechanisms, which will be reported separately. Finally, optimal task weighting and architecture settings ($\lambda_{y,s,r}, d, \tau$) are likely to be task-specific. We expect an exciting direction for future work will be in, e.g., Bayesian optimization of these parameters.

%We note there are still major challenges in constructing maximally transferrable pre-training sets as well as in modeling at the pre-training scale. Studies of our approach in the self-supervised setting are ongoing and will be reported in the near future.

Overall, we anticipate the findings herein to aid in the training of more robust 3D GNNs on molecular conformers. We expect the approach to be applicable in many 3D modeling tasks for small and large molecules, e.g., proteins. SupSiam may find particular utility in settings where sensitivity to minor 2D structure changes, but insensitivity to minor 3D structure changes, are desirable. We are hopeful it will aid in overcoming challenges such as activity cliffs \citep{Stumpfe} and rough SAR landscapes \citep{Aldeghi}, and will lead to more reliable modeling in MLDD.

%%%%%%%%%%%%%%%%%%%%%%%%%%%%%%%%%%%%%%%%%%%%%%%%%%%%%%%%%
\bibliography{main}

\begin{thebibliography}{30}
\providecommand{\natexlab}[1]{#1}
\providecommand{\url}[1]{\texttt{#1}}
\expandafter\ifx\csname urlstyle\endcsname\relax
  \providecommand{\doi}[1]{doi: #1}\else
  \providecommand{\doi}{doi: \begingroup \urlstyle{rm}\Url}\fi

\bibitem[Aldeghi et~al.(2022)Aldeghi, Graff, Frey, Morrone, Pyzer-Knapp,
  Jordan, and Coley]{Aldeghi}
Matteo Aldeghi, David~E. Graff, Nathan Frey, Joseph~A. Morrone, Edward~O.
  Pyzer-Knapp, Kirk~E. Jordan, and Connor~W. Coley.
\newblock {Roughness of Molecular Property Landscapes and Its Impact on
  Modellability}.
\newblock \emph{Journal of Chemical Information and Modeling}, 62\penalty0
  (19):\penalty0 4660--4671, 2022.
\newblock ISSN 1549-9596.
\newblock \doi{10.1021/acs.jcim.2c00903}.

\bibitem[Axelrod \& Gomez-Bombarelli(2020)Axelrod and
  Gomez-Bombarelli]{Axelroddqd}
Simon Axelrod and Rafael Gomez-Bombarelli.
\newblock {Molecular machine learning with conformer ensembles}.
\newblock \emph{arXiv}, 2020.
\newblock \doi{10.48550/arxiv.2012.08452}.

\bibitem[Batatia et~al.(2022)Batatia, Kovács, Simm, Ortner, and
  Csányi]{Batatia}
Ilyes Batatia, Dávid~Péter Kovács, Gregor N~C Simm, Christoph Ortner, and
  Gábor Csányi.
\newblock {MACE: Higher Order Equivariant Message Passing Neural Networks for
  Fast and Accurate Force Fields}.
\newblock \emph{arXiv}, 2022.
\newblock \doi{10.48550/arxiv.2206.07697}.

\bibitem[Chen \& He(2020)Chen and He]{Chen}
Xinlei Chen and Kaiming He.
\newblock {Exploring Simple Siamese Representation Learning}.
\newblock \emph{arXiv}, 2020.
\newblock \doi{10.48550/arxiv.2011.10566}.

\bibitem[Devereux et~al.(2020)Devereux, Smith, Davis, Barros, Zubatyuk, Isayev,
  and Roitberg]{Devereux}
Christian Devereux, Justin~S. Smith, Kate~K. Davis, Kipton Barros, Roman
  Zubatyuk, Olexandr Isayev, and Adrian~E. Roitberg.
\newblock {Extending the Applicability of the ANI Deep Learning Molecular
  Potential to Sulfur and Halogens}.
\newblock \emph{Journal of Chemical Theory and Computation}, 16\penalty0
  (7):\penalty0 4192--4202, 2020.
\newblock ISSN 1549-9618.
\newblock \doi{10.1021/acs.jctc.0c00121}.

\bibitem[Ganea et~al.(2021{\natexlab{a}})Ganea, Huang, Bunne, Bian, Barzilay,
  Jaakkola, and Krause]{Ganearks}
Octavian-Eugen Ganea, Xinyuan Huang, Charlotte Bunne, Yatao Bian, Regina
  Barzilay, Tommi Jaakkola, and Andreas Krause.
\newblock {Independent SE(3)-Equivariant Models for End-to-End Rigid Protein
  Docking}.
\newblock \emph{arXiv}, 2021{\natexlab{a}}.
\newblock \doi{10.48550/arxiv.2111.07786}.

\bibitem[Ganea et~al.(2021{\natexlab{b}})Ganea, Pattanaik, Coley, Barzilay,
  Jensen, Green, and Jaakkola]{Ganea}
Octavian-Eugen Ganea, Lagnajit Pattanaik, Connor~W Coley, Regina Barzilay,
  Klavs~F Jensen, William~H Green, and Tommi~S Jaakkola.
\newblock {GeoMol: Torsional Geometric Generation of Molecular 3D Conformer
  Ensembles}.
\newblock \emph{arXiv}, 2021{\natexlab{b}}.
\newblock \doi{10.48550/arxiv.2106.07802}.

\bibitem[Geiger \& Smidt(2022)Geiger and Smidt]{Geiger}
Mario Geiger and Tess Smidt.
\newblock {e3nn: Euclidean Neural Networks}.
\newblock \emph{arXiv}, 2022.
\newblock \doi{10.48550/arxiv.2207.09453}.

\bibitem[Godwin et~al.(2021)Godwin, Schaarschmidt, Gaunt, Sanchez-Gonzalez,
  Rubanova, Veličković, Kirkpatrick, and Battaglia]{Godwin}
Jonathan Godwin, Michael Schaarschmidt, Alexander Gaunt, Alvaro
  Sanchez-Gonzalez, Yulia Rubanova, Petar Veličković, James Kirkpatrick, and
  Peter Battaglia.
\newblock {Simple GNN Regularisation for 3D Molecular Property Prediction \&
  Beyond}.
\newblock \emph{arXiv}, 2021.
\newblock \doi{10.48550/arxiv.2106.07971}.

\bibitem[Grill et~al.(2020)Grill, Strub, Altché, Tallec, Richemond,
  Buchatskaya, Doersch, Pires, Guo, Azar, Piot, Kavukcuoglu, Munos, and
  Valko]{Grill}
Jean-Bastien Grill, Florian Strub, Florent Altché, Corentin Tallec, Pierre~H
  Richemond, Elena Buchatskaya, Carl Doersch, Bernardo~Avila Pires,
  Zhaohan~Daniel Guo, Mohammad~Gheshlaghi Azar, Bilal Piot, Koray Kavukcuoglu,
  Rémi Munos, and Michal Valko.
\newblock {Bootstrap your own latent: A new approach to self-supervised
  Learning}.
\newblock \emph{arXiv}, 2020.
\newblock \doi{10.48550/arxiv.2006.07733}.

\bibitem[Huang et~al.(2022)Huang, Fu, Gao, Zhao, Roohani, Leskovec, Coley,
  Xiao, Sun, and Zitnik]{Huang}
Kexin Huang, Tianfan Fu, Wenhao Gao, Yue Zhao, Yusuf Roohani, Jure Leskovec,
  Connor~W. Coley, Cao Xiao, Jimeng Sun, and Marinka Zitnik.
\newblock {Artificial intelligence foundation for therapeutic science}.
\newblock \emph{Nature Chemical Biology}, 18\penalty0 (10):\penalty0
  1033--1036, 2022.
\newblock ISSN 1552-4450.
\newblock \doi{10.1038/s41589-022-01131-2}.

\bibitem[Isert et~al.(2022)Isert, Atz, Jiménez-Luna, and Schneider]{Isert}
Clemens Isert, Kenneth Atz, José Jiménez-Luna, and Gisbert Schneider.
\newblock {QMugs, quantum mechanical properties of drug-like molecules}.
\newblock \emph{Scientific Data}, 9\penalty0 (1):\penalty0 273, 2022.
\newblock \doi{10.1038/s41597-022-01390-7}.

\bibitem[Jones et~al.(2021)Jones, Kim, Zhang, Zemla, Stevenson, Bennett,
  Kirshner, Wong, Lightstone, and Allen]{Jones}
Derek Jones, Hyojin Kim, Xiaohua Zhang, Adam Zemla, Garrett Stevenson,
  W.~F.~Drew Bennett, Daniel Kirshner, Sergio~E. Wong, Felice~C. Lightstone,
  and Jonathan~E. Allen.
\newblock {Improved Protein–Ligand Binding Affinity Prediction with
  Structure-Based Deep Fusion Inference}.
\newblock \emph{Journal of Chemical Information and Modeling}, 61\penalty0
  (4):\penalty0 1583--1592, 2021.
\newblock ISSN 1549-9596.
\newblock \doi{10.1021/acs.jcim.0c01306}.

\bibitem[Kendall \& Gal(2017)Kendall and Gal]{Kendall}
Alex Kendall and Yarin Gal.
\newblock {What Uncertainties Do We Need in Bayesian Deep Learning for Computer
  Vision?}
\newblock \emph{arXiv}, 2017.
\newblock \doi{10.48550/arxiv.1703.04977}.

\bibitem[Kohn \& Sham(1965)Kohn and Sham]{Kohn}
W.~Kohn and L.~J. Sham.
\newblock {Self-Consistent Equations Including Exchange and Correlation
  Effects}.
\newblock \emph{Physical Review}, 140\penalty0 (4A):\penalty0 A1133--A1138,
  1965.
\newblock ISSN 0031-899X.
\newblock \doi{10.1103/physrev.140.a1133}.

\bibitem[Kombo et~al.(2013)Kombo, Tallapragada, Jain, Chewning, Mazurov,
  Speake, Hauser, and Toler]{Kombo}
David~C. Kombo, Kartik Tallapragada, Rachit Jain, Joseph Chewning, Anatoly~A.
  Mazurov, Jason~D. Speake, Terry~A. Hauser, and Steve Toler.
\newblock {3D Molecular Descriptors Important for Clinical Success}.
\newblock \emph{Journal of Chemical Information and Modeling}, 53\penalty0
  (2):\penalty0 327--342, 2013.
\newblock ISSN 1549-9596.
\newblock \doi{10.1021/ci300445e}.

\bibitem[Landrum(2022)]{ny}
Greg Landrum.
\newblock {RDKit: Open-source cheminformatics.}, 3 2022.
\newblock URL \url{https://www.rdkit.org}.

\bibitem[Le-Khac et~al.(2020)Le-Khac, Healy, and Smeaton]{Le-Khac}
Phuc~H. Le-Khac, Graham Healy, and Alan~F. Smeaton.
\newblock {Contrastive Representation Learning: A Framework and Review}.
\newblock \emph{IEEE Access}, 8:\penalty0 193907--193934, 2020.
\newblock ISSN 2169-3536.
\newblock \doi{10.1109/access.2020.3031549}.

\bibitem[Li et~al.(2022)Li, Efros, and Pathak]{Li}
Alexander~C Li, Alexei~A Efros, and Deepak Pathak.
\newblock {Understanding Collapse in Non-Contrastive Siamese Representation
  Learning}.
\newblock \emph{arXiv}, 2022.
\newblock \doi{10.48550/arxiv.2209.15007}.

\bibitem[Liao \& Smidt(2022)Liao and Smidt]{Liao}
Yi-Lun Liao and Tess Smidt.
\newblock {Equiformer: Equivariant Graph Attention Transformer for 3D Atomistic
  Graphs}.
\newblock \emph{arXiv}, 2022.
\newblock \doi{10.48550/arxiv.2206.11990}.

\bibitem[Ng et~al.(2022)Ng, Hulkund, Cho, and Ghassemi]{Ng}
Nathan Ng, Neha Hulkund, Kyunghyun Cho, and Marzyeh Ghassemi.
\newblock {Predicting Out-of-Domain Generalization with Local Manifold
  Smoothness}.
\newblock \emph{arXiv}, 2022.
\newblock \doi{10.48550/arxiv.2207.02093}.

\bibitem[Rackers et~al.(2022)Rackers, Tecot, Geiger, and Smidt]{Rackers}
Joshua~A Rackers, Lucas Tecot, Mario Geiger, and Tess~E Smidt.
\newblock {Cracking the Quantum Scaling Limit with Machine Learned Electron
  Densities}.
\newblock \emph{arXiv}, 2022.
\newblock \doi{10.48550/arxiv.2201.03726}.

\bibitem[Sauer \& Schwarz(2003)Sauer and Schwarz]{Sauer}
Wolfgang H.~B. Sauer and Matthias~K. Schwarz.
\newblock {Molecular Shape Diversity of Combinatorial Libraries: A Prerequisite
  for Broad Bioactivity †}.
\newblock \emph{Journal of Chemical Information and Computer Sciences},
  43\penalty0 (3):\penalty0 987--1003, 2003.
\newblock ISSN 0095-2338.
\newblock \doi{10.1021/ci025599w}.

\bibitem[Stumpfe et~al.(2019)Stumpfe, Hu, and Bajorath]{Stumpfe}
Dagmar Stumpfe, Huabin Hu, and Juergen Bajorath.
\newblock {Evolving Concept of Activity Cliffs}.
\newblock \emph{ACS Omega}, 4\penalty0 (11):\penalty0 14360--14368, 2019.
\newblock ISSN 2470-1343.
\newblock \doi{10.1021/acsomega.9b02221}.

\bibitem[Thomas et~al.(2018)Thomas, Smidt, Kearnes, Yang, Li, Kohlhoff, and
  Riley]{Thomas}
Nathaniel Thomas, Tess Smidt, Steven Kearnes, Lusann Yang, Li~Li, Kai Kohlhoff,
  and Patrick Riley.
\newblock {Tensor field networks: Rotation- and translation-equivariant neural
  networks for 3D point clouds}.
\newblock \emph{arXiv}, 2018.
\newblock \doi{10.48550/arxiv.1802.08219}.

\bibitem[Wang et~al.(2020)Wang, Witek, Landrum, and Riniker]{Wang}
Shuzhe Wang, Jagna Witek, Gregory~A. Landrum, and Sereina Riniker.
\newblock {Improving Conformer Generation for Small Rings and Macrocycles Based
  on Distance Geometry and Experimental Torsional-Angle Preferences}.
\newblock \emph{Journal of Chemical Information and Modeling}, 60\penalty0
  (4):\penalty0 2044--2058, 2020.
\newblock ISSN 1549-9596.
\newblock \doi{10.1021/acs.jcim.0c00025}.

\bibitem[Wang et~al.(2021)Wang, Wang, Cao, and Farimani]{Wangynb}
Yuyang Wang, Jianren Wang, Zhonglin Cao, and Amir~Barati Farimani.
\newblock {Molecular Contrastive Learning of Representations via Graph Neural
  Networks}.
\newblock \emph{arXiv}, 2021.
\newblock \doi{10.48550/arxiv.2102.10056}.

\bibitem[Wu et~al.(2017)Wu, Ramsundar, Feinberg, Gomes, Geniesse, Pappu,
  Leswing, and Pande]{Wu}
Zhenqin Wu, Bharath Ramsundar, Evan~N. Feinberg, Joseph Gomes, Caleb Geniesse,
  Aneesh~S. Pappu, Karl Leswing, and Vijay Pande.
\newblock {MoleculeNet: a benchmark for molecular machine learning}.
\newblock \emph{Chemical Science}, 9\penalty0 (2):\penalty0 513--530, 2017.
\newblock ISSN 2041-6520.
\newblock \doi{10.1039/c7sc02664a}.

\bibitem[Zaidi et~al.(2022)Zaidi, Schaarschmidt, Martens, Kim, Teh,
  Sanchez-Gonzalez, Battaglia, Pascanu, and Godwin]{Zaidi}
Sheheryar Zaidi, Michael Schaarschmidt, James Martens, Hyunjik Kim, Yee~Whye
  Teh, Alvaro Sanchez-Gonzalez, Peter Battaglia, Razvan Pascanu, and Jonathan
  Godwin.
\newblock {Pre-training via Denoising for Molecular Property Prediction}.
\newblock \emph{arXiv}, 2022.
\newblock \doi{10.48550/arxiv.2206.00133}.

\bibitem[Zheng et~al.(2017)Zheng, Tice, and Singh]{Zheng}
Yajun Zheng, Colin~M. Tice, and Suresh~B. Singh.
\newblock {Conformational control in structure-based drug design}.
\newblock \emph{Bioorganic \& Medicinal Chemistry Letters}, 27\penalty0
  (13):\penalty0 2825--2837, 2017.
\newblock ISSN 0960-894X.
\newblock \doi{10.1016/j.bmcl.2017.04.079}.

\end{thebibliography}
\bibliographystyle{iclr2023_conference}
%%%%%%%%%%%%%%%%%%%%%%%%%%%%%%%%%%%%%%%%%%%%%%%%%%%%%%%%%

%%%%%%%%%%%%%%%%%%%%%%%%%%%%%%%%%%%%%%%%%%%%%%%%%%%%%%%%%%%%

% For faster compiling
%\end{document}

\appendix

\section{Appendix}

\subsection{Data processing}

\subsubsection{Target distributions}

See Section \ref{sec:datasets} for high-level data-processing details. For the \textbf{}{Clear} task, targets were scaled by $log_10$ to give a roughly Gaussian prior. For all datasets, oversampling was utilized, where model batches were sampled from the training set weighted proportionally to the inverse of their histogram-bin density as follows:

\begin{equation}
    w_i = 1 - p_y(y_i) ,
\end{equation}

where $p_y$ represents the prior density distribution.

\subsubsection{Conformer ensemble generation}

All processing functions are made available in the associated code repository at \href{}{[link to be activated on acceptance for publication]}. The conformer generation pipeline follows that in \citet{Axelroddqd} closely. A model pipeline is as follows:

\begin{algorithm}[H] 
\caption{Conformer ensemble generation}\label{alg:ce}
\begin{algorithmic}[1]

\State{{\textbf{Input}:}} Dataset of $N$ observations $\mathcal{D}_n$, with input space $\mathcal{X}$ as molecular SMILES strings, conformer ensemble size $c$, boolean optimize $o$, boolean align $a$, \texttt{rdkit.Chem.AllChem} package, \texttt{rdkit.Chem.rdMolAlign} package \\ 

\textbf{Output:} Dataset of $N$ observations $\mathcal{D}_n$, with input space atomic positions $\mathcal{X} \in \mathbb{R}^3$.

\For{$\lbrace n = 1, \dots, N \rbrace$}

\State{Construct \texttt{AllChem.Mol} object $\mathcal{M}_n$}

\State{$\mathcal{M}_n \leftarrow$  \texttt{AllChem.AddHs(AllChem.MolFromSmiles($\mathcal{X}_n$))}}

\State{Embed ensemble of 3D conformers with ETKDG}
\State{$\mathcal{M}^C_n \leftarrow$ \texttt{AllChem.EmbedMultipleConfs($\mathcal{M}_n$, numConfs=$c$)}}

\If{$o$}

\State{Optimize molecular conformers with force field}
\State{$\mathcal{M}^C_n \leftarrow$ \texttt{AllChem.UFFOptimizeMoleculeConfs($\mathcal{M}^C_n$, numConfs=$c$)}}

\EndIf

\State{Remove hydrogen atoms}
\State{$\mathcal{M}_n \leftarrow$ \texttt{AllChem.RemoveHs($\mathcal{M}_n$)}}

\If{$a$}

\State{Align molecular conformers}
\State{$\mathcal{M}^C_n \leftarrow$ \texttt{rdMolAlign.AlignMolConformers($\mathcal{M}^C_n$)}}

\EndIf

\EndFor

\end{algorithmic}
\end{algorithm}

\subsection{Model architecture}

\subsubsection{E3NNs}

See Section \ref{sec:siamese_e3nns} for high-level architecture details. Full hyperparameters are included below for a network with hidden dimension $d = 128$:

\begin{table}[H]
  \caption{E3NN hyperparameter settings}
  \label{tab:e3nn}
  \centering
  \begin{tabular}{ccc}
    \toprule
    %\multicolumn{2}{c}{Part}                   \\
    %\cmidrule(r){1-2}
    Name     & Setting     & Description \\
    \midrule
    irreps\_in & 128x0e  & input-layer irreducible representations     \\
    irreps\_hidden & 128x0e+128x1o+x2e  & hidden-layer irreducible representations     \\
    irreps\_out & 128x0e  & output-layer irreducible representations     \\
    l\_max\_sh & 1  & maximum geometric tensor spherical-harmonic level    \\
    num\_hidden\_layers & 4  & number of E3NN convolution layers    \\
    rc & 4.0\AA  & neighborhood-edge radial cutoff distance  \\
    irreps\_edge & 128x0e  & edge-layer irreducible representations     \\
    radial\_num\_basis & 16  & number of basis functions for radial NN     \\
    radial\_num\_hidden  & 16  & radial NN hidden dimension    \\
    radial\_num\_layers & 2  & radial NN depth     \\
    add\_self\_loops & True  & include self-edges in radial graphs     \\
    \bottomrule
  \end{tabular}
\end{table}

Intermediate layers are treated with ShiftedSoftplus activation and \texttt{LayerNorm}.

E3NNs output feature vectors for each node, comprising their flattened and concatenated geometric tensors. These embeddings are pooled by \texttt{global\_mean\_pool} to give single vector representations for each molecule, to which a final linear readout layer is applied.

\subsubsection{Siamese projection MLP}

For the Siamese task, E3NN readout representations of dimension $d$ are input to a multilayer perceptron (MLP) of depth 2 and dimensions \{$d \times 2$, $d$\}. Both layers are followed by ShiftedSoftplus activation, and the intermediate representations are treated with \texttt{LayerNorm} and dropout of probability 0.2. 

\subsubsection{Probabilistic MLP}

The probabilistic predictive model is a split-head MLP with two modules, each of depth 3 and output dimensions \{$d\times 2$, $d$, 1\}. Intermediate layers are followed by  ShiftedSoftplus activation, \texttt{LayerNorm}, and dropout with probability 0.2. The mean ($\mu$) module is unactivated, outputting raw logit values. The variance ($\sigma$) module outputs are activated by Softplus to give predicted posterior distributions $(\mu, \sigma)$. Following \citet{Kendall}, $m$ samples are drawn from these distributions, and the resulting logits are activated by \texttt{sigmoid} for classification tasks and \texttt{Tanhshrink} for regression tasks. In the case of regression, the resulting predictions are scaled using the prior parameters $(\mu_t, \sigma_t)$ computed on the training set. An aggregate loss is computed over the sampled predictions, binary cross entropy (BCE) for classification and mean squared error (MSE) for regression.

\subsection{Full results}\label{app:sec:results}

A full grid search study was run over the following hyperparameter ranges:

\begin{table}[H]
  \caption{SupSiam hyperparameter screen}
  \label{tab:ablation}
  \centering
  \begin{tabular}{ccc}
    \toprule
    Name     & Values     & Description \\
    \midrule
    $n$ & 10  & number of model runs     \\
    $e$ & 50  & number of training epochs     \\
    $\tau$ & [0.1, 1]  & node noise multiplier     \\
    $d$ & [128, 256]  & hidden dimension    \\
    $\lambda_s$ & [0.0, 0.1, 1.0, 10.0]  & Siamese loss weight     \\
    $\lambda_r$ & [0.0, 0.1, 1.0, 10.0]  & $l_2$ loss weight     \\
    \bottomrule
  \end{tabular}
\end{table}

\subsubsection{Training profiles}\label{app:sec:training}

Model performance and properties were tracked throughout training and included $\mathcal{L}_y$, $\mathcal{L}_s$, $\mathcal{L}_r$, and $\sigma_z^2$ (see Equations \ref{eqn:loss}, \ref{eqn:cos}). \texttt{roc\_auc\_score}s were additionally tracked for classification tasks, and Spearman $\rho$ for regression.

Full results are provided below. Each figure contains 1 metric above, from 1 task, and at 1 value of $\tau$ over the course of training. Plots are arranged with increasing $d$ along the horizontal axis, and increasing $\lambda_r$ along the vertical axis. Scatter plots like Figure \ref{fig:rocauc} are also included, where all models were loaded from the epoch of their best validation set metric performance (\texttt{roc\_auc\_score} or Spearman $\rho$). These are organized with increasing $d$ along the horizontal, increasing $\tau$ along the vertical, and increasing $\lambda_r$ on the subplot $x$-axis. 

Note that for the \textbf{Pgp} task, \texttt{roc\_auc\_score}s in training figures were calculated with predictions binarized at $\hat{y} \geq 0.99$. For the scatter plots, however, full ROC curves were plotted over a range of 100 binarization thresholds evenly spaced from [0.0, 1.0]. Areas under these curves were then directly computed for each repeat model, giving the results shown in the plots.

\clearpage

\subsubsection{\textbf{Pgp}}

\begin{figure}[H]
    \centering
    \includegraphics[width=0.98\textwidth]{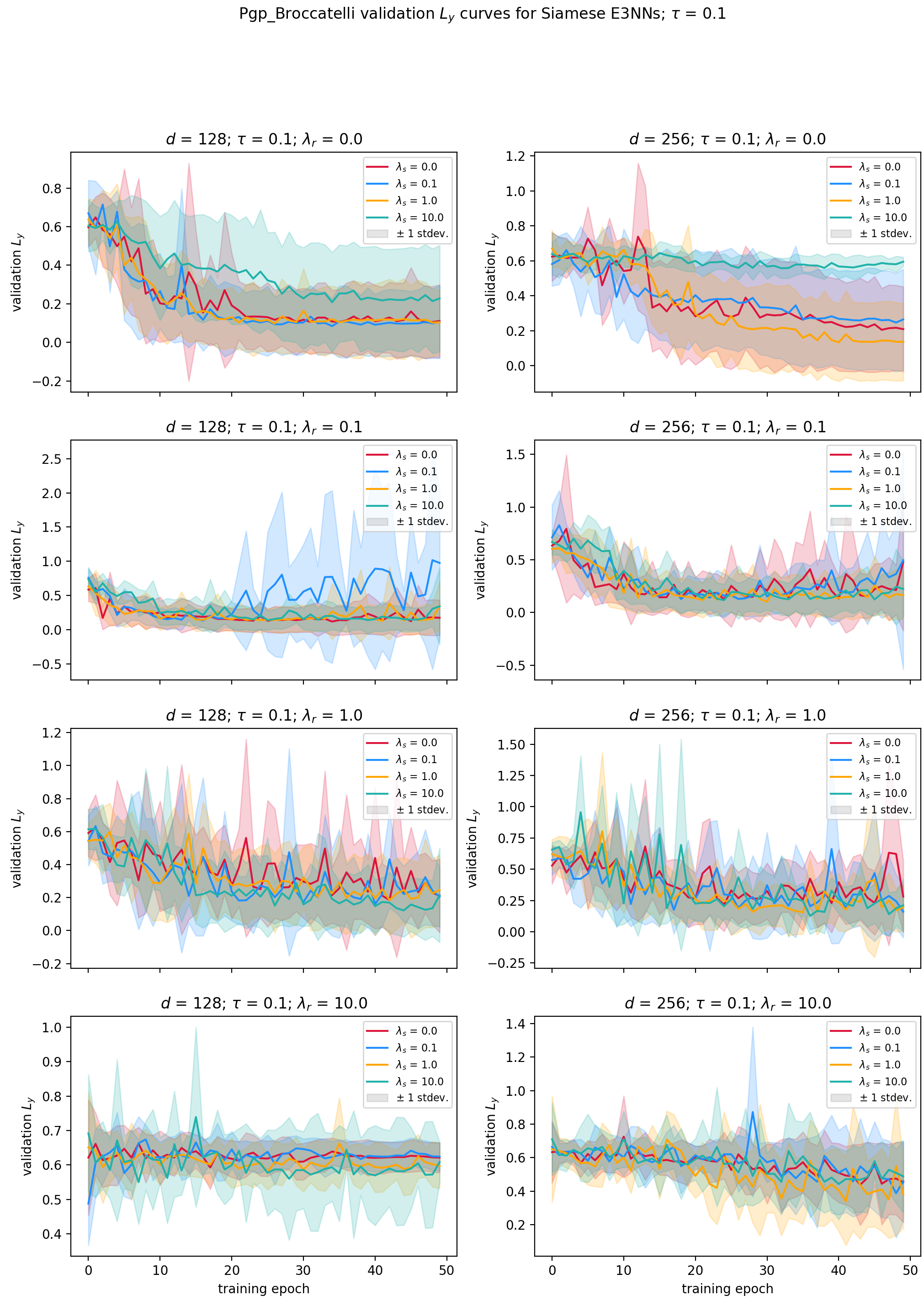}
    %\caption{}
    \label{app:fig:pgp_ly_0.1}
\end{figure}

\begin{figure}[t]
    \centering
    \includegraphics[width=0.98\textwidth]{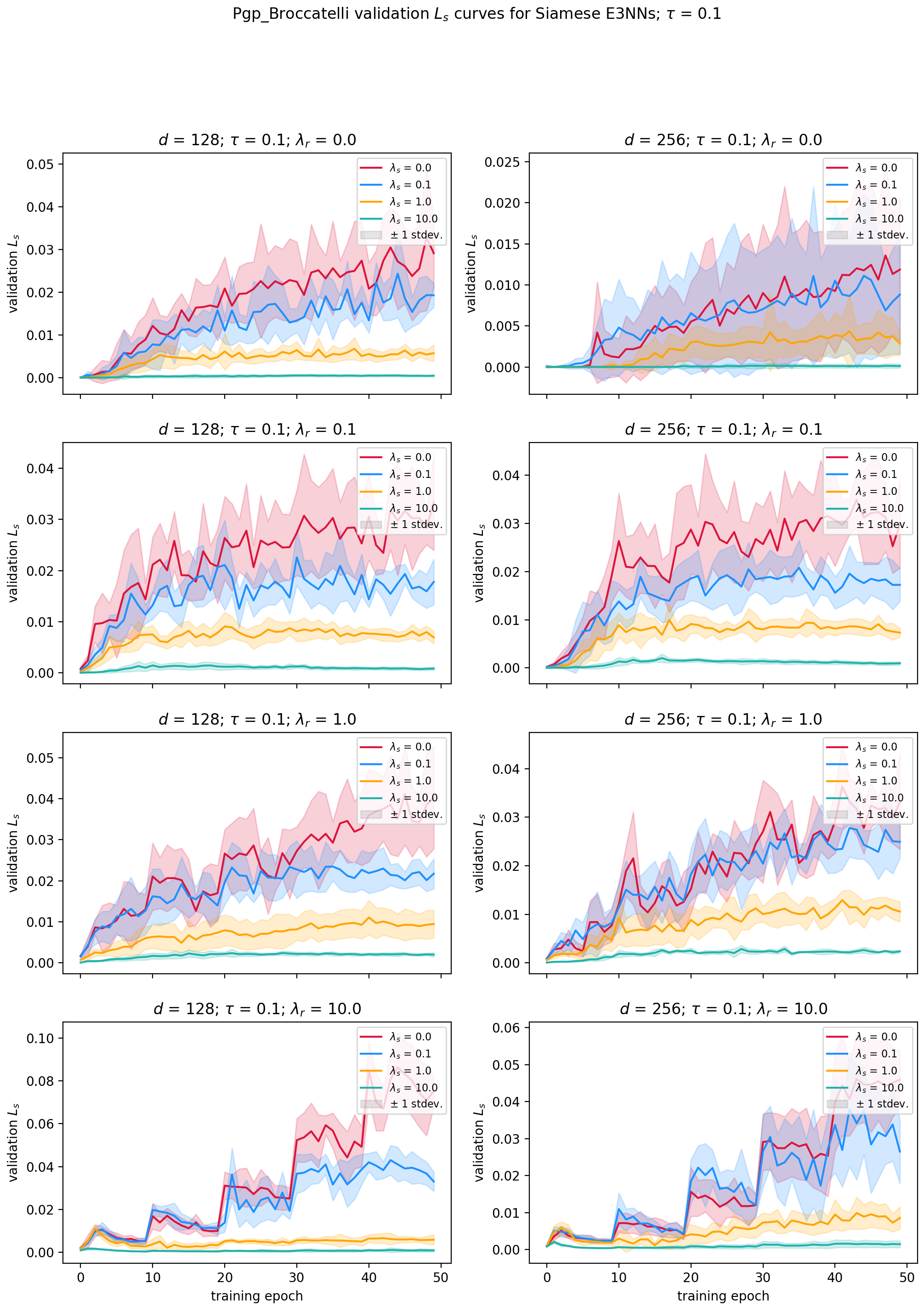}
    %\caption{}
    \label{app:fig:pgp_ls_0.1}
\end{figure}

\begin{figure}[t]
    \centering
    \includegraphics[width=0.98\textwidth]{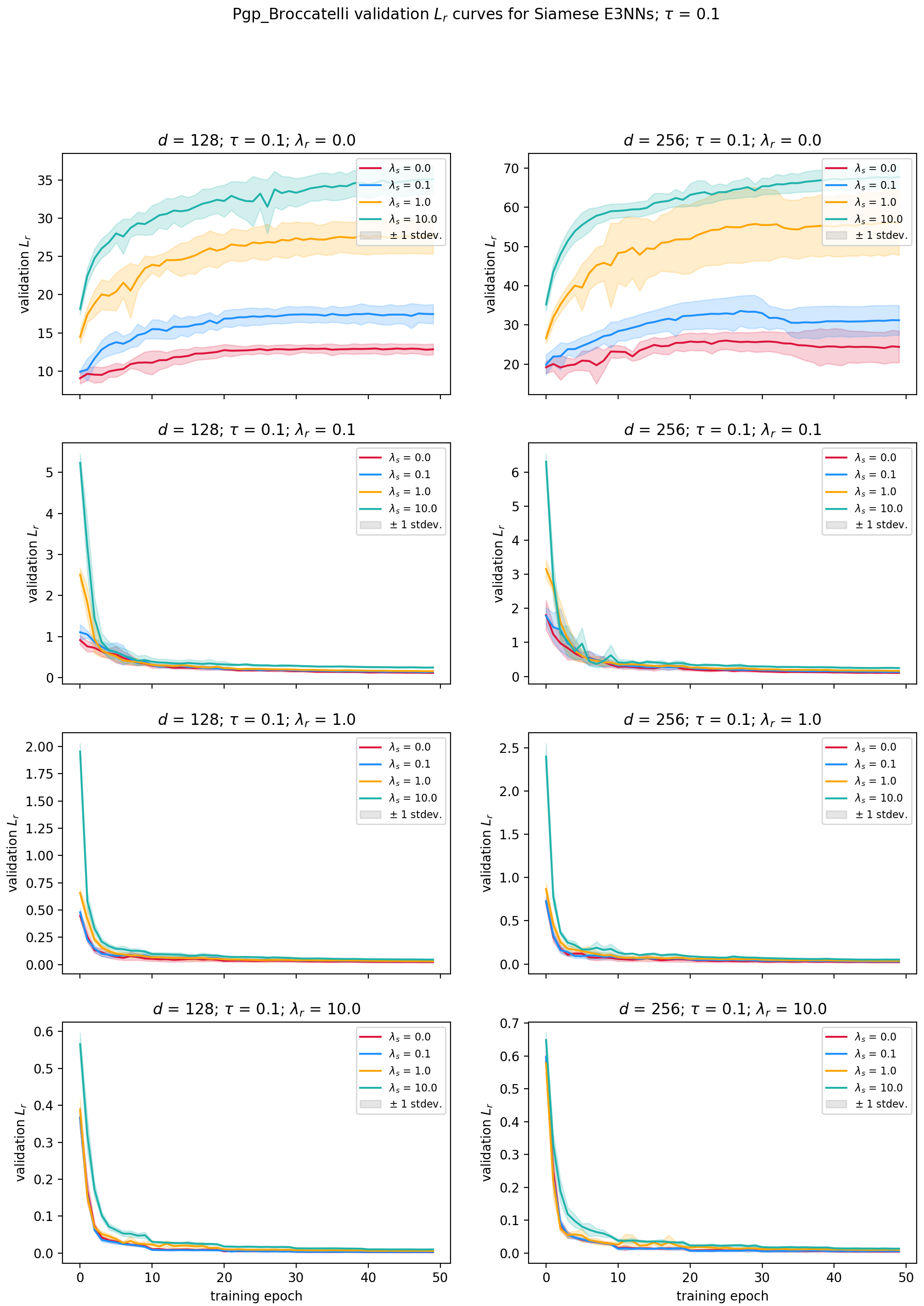}
    %\caption{}
    \label{app:fig:pgp_lr_0.1}
\end{figure}

\begin{figure}[t]
    \centering
    \includegraphics[width=0.98\textwidth]{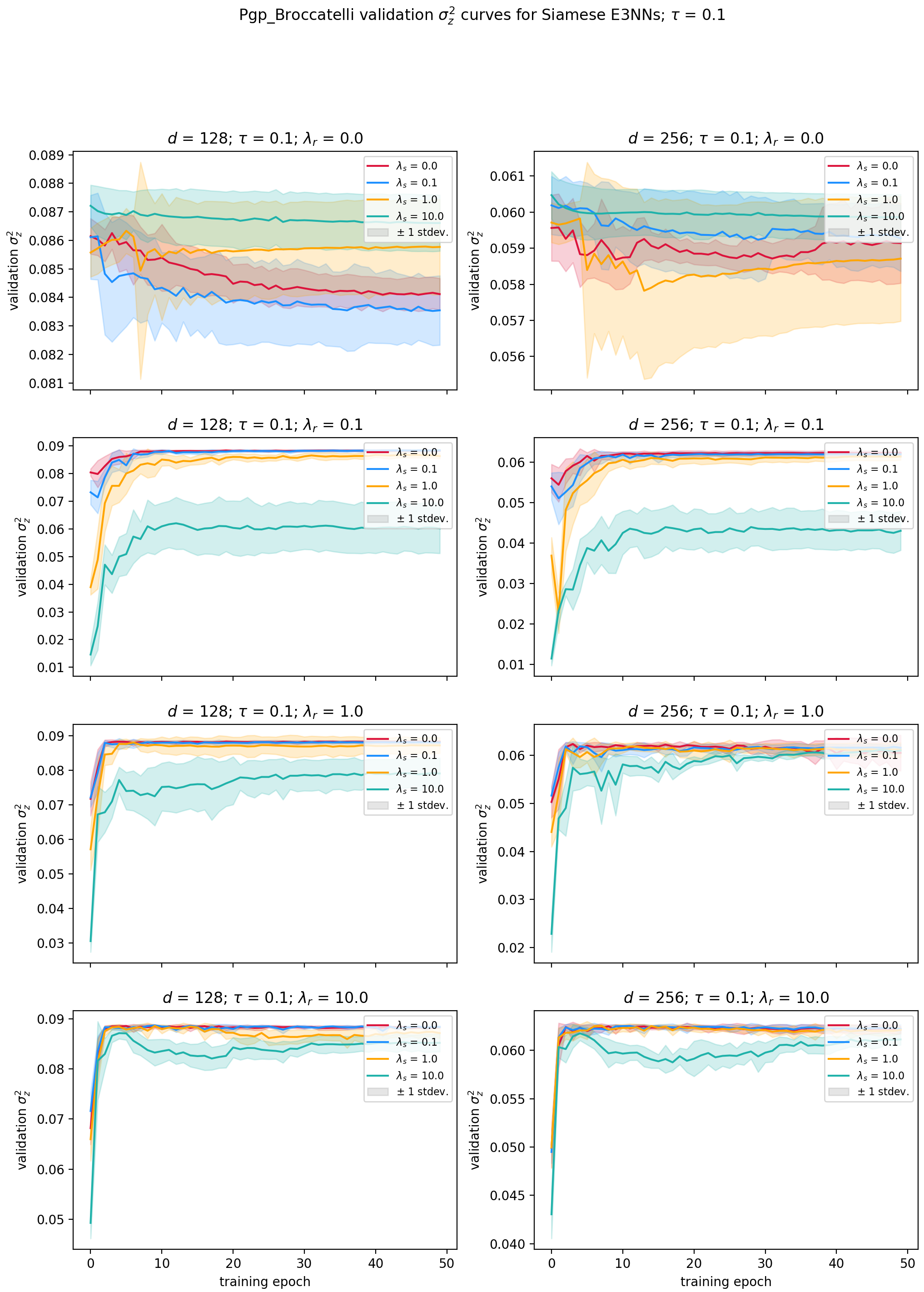}
    %\caption{}
    \label{app:fig:pgp_sig_0.1}
\end{figure}

\begin{figure}[t]
    \centering
    \includegraphics[width=0.98\textwidth]{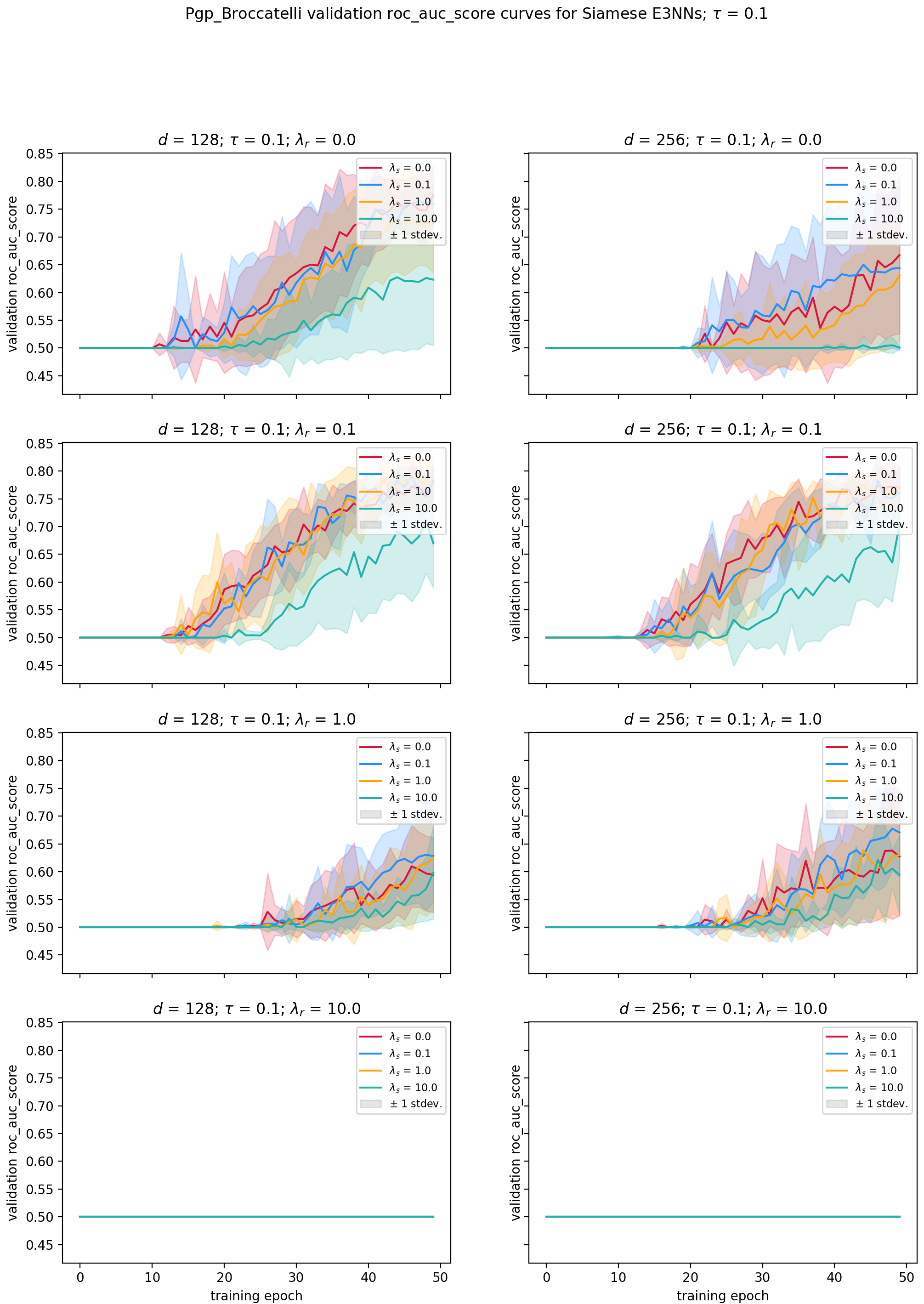}
    %\caption{}
    \label{app:fig:pgp_roc_0.1}
\end{figure}

\begin{figure}[t]
    \centering
    \includegraphics[width=0.98\textwidth]{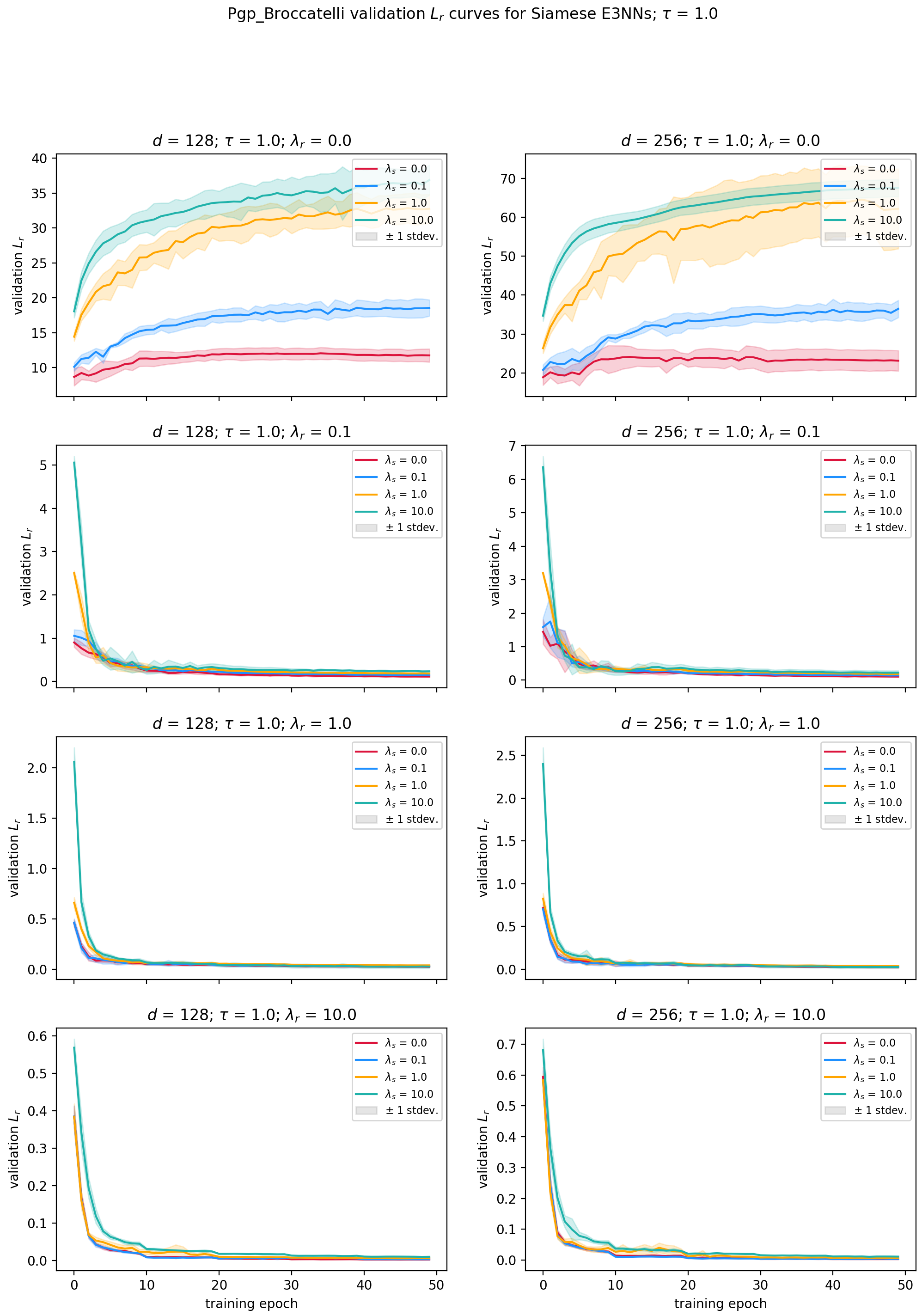}
    %\caption{}
    \label{app:fig:pgp_ly_1.0}
\end{figure}

\begin{figure}[t]
    \centering
    \includegraphics[width=0.98\textwidth]{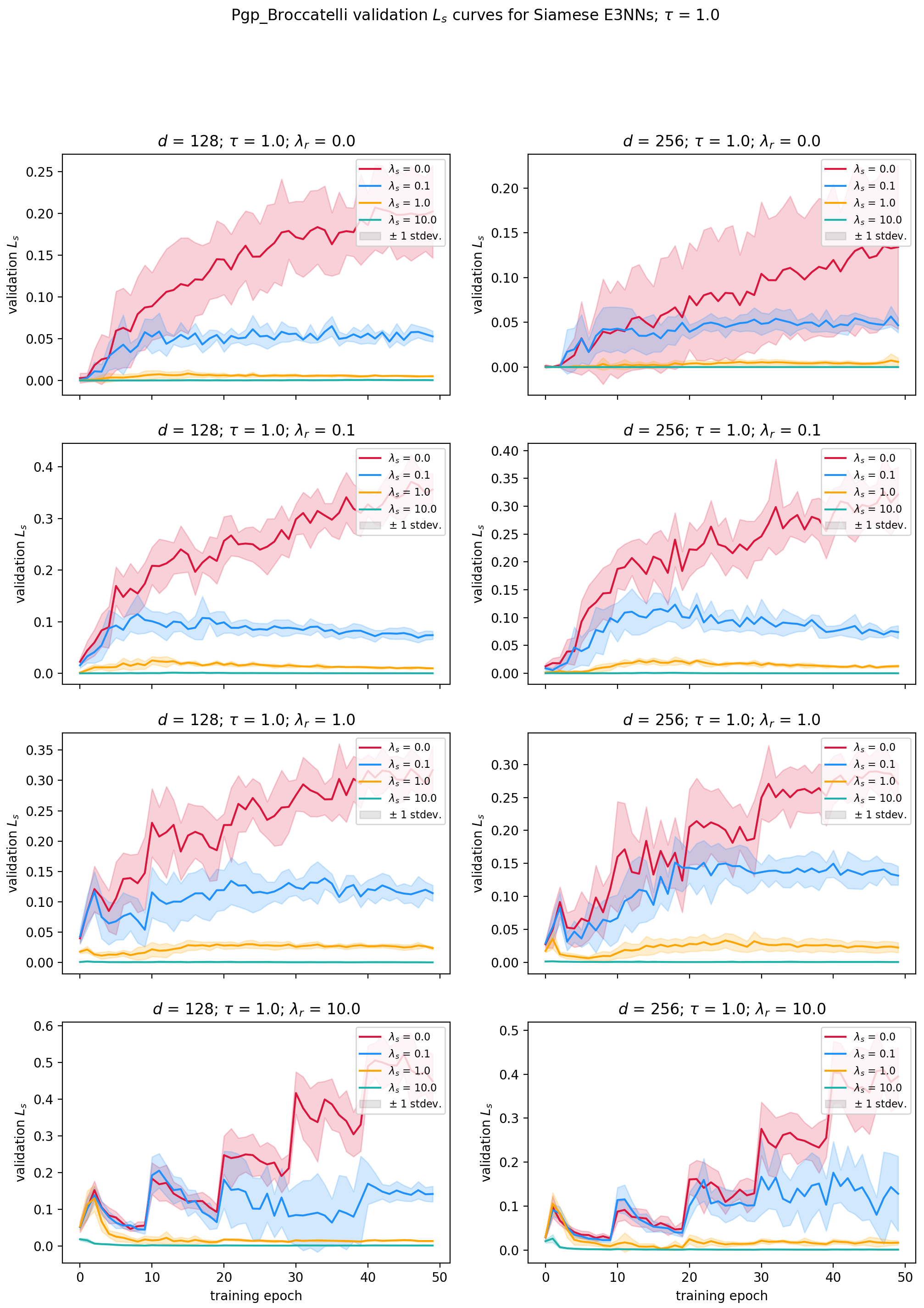}
    %\caption{}
    \label{app:fig:pgp_ls_1.0}
\end{figure}

\begin{figure}[t]
    \centering
    \includegraphics[width=0.98\textwidth]{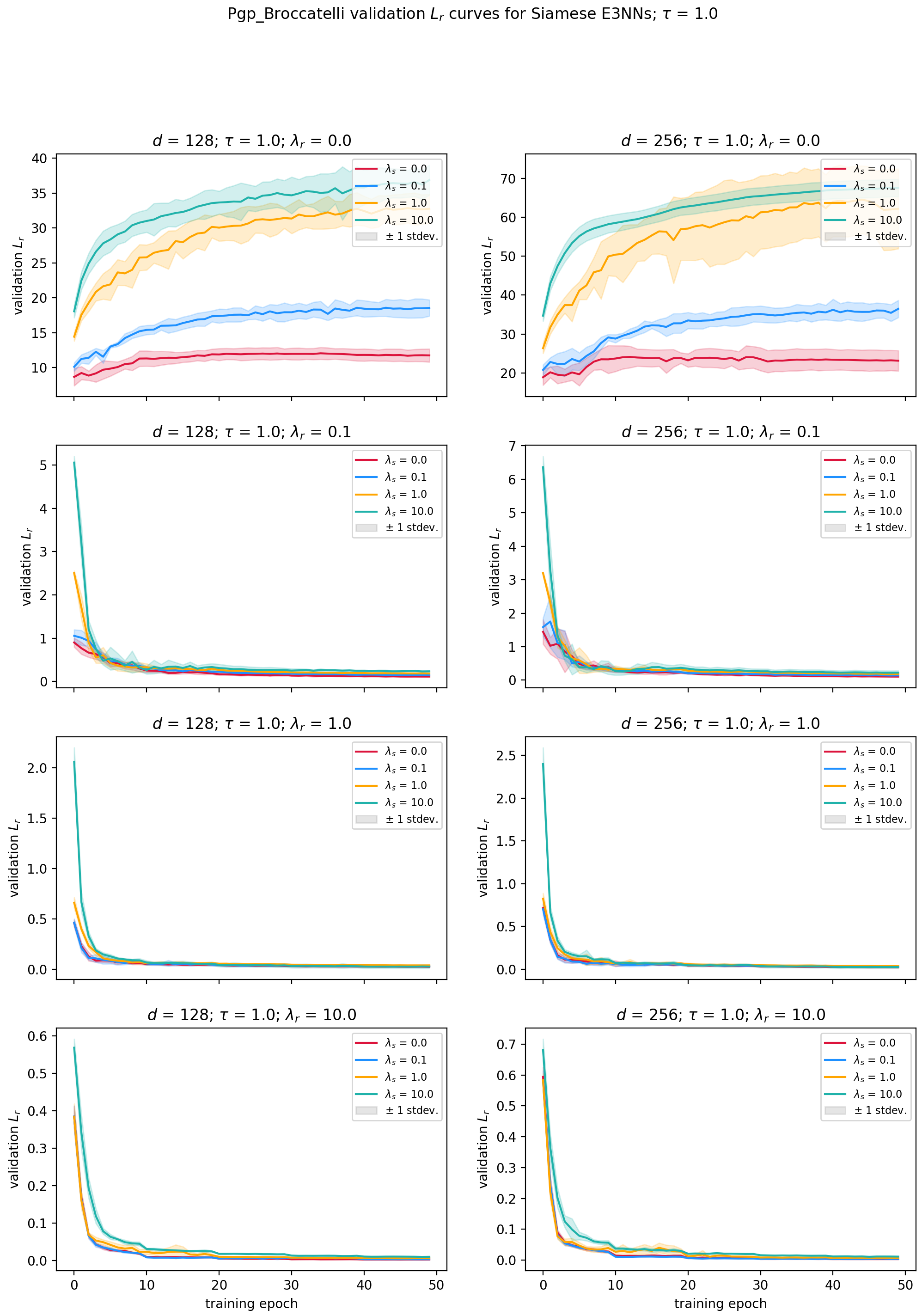}
    %\caption{}
    \label{app:fig:pgp_lr_1.0}
\end{figure}

\begin{figure}[t]
    \centering
    \includegraphics[width=0.98\textwidth]{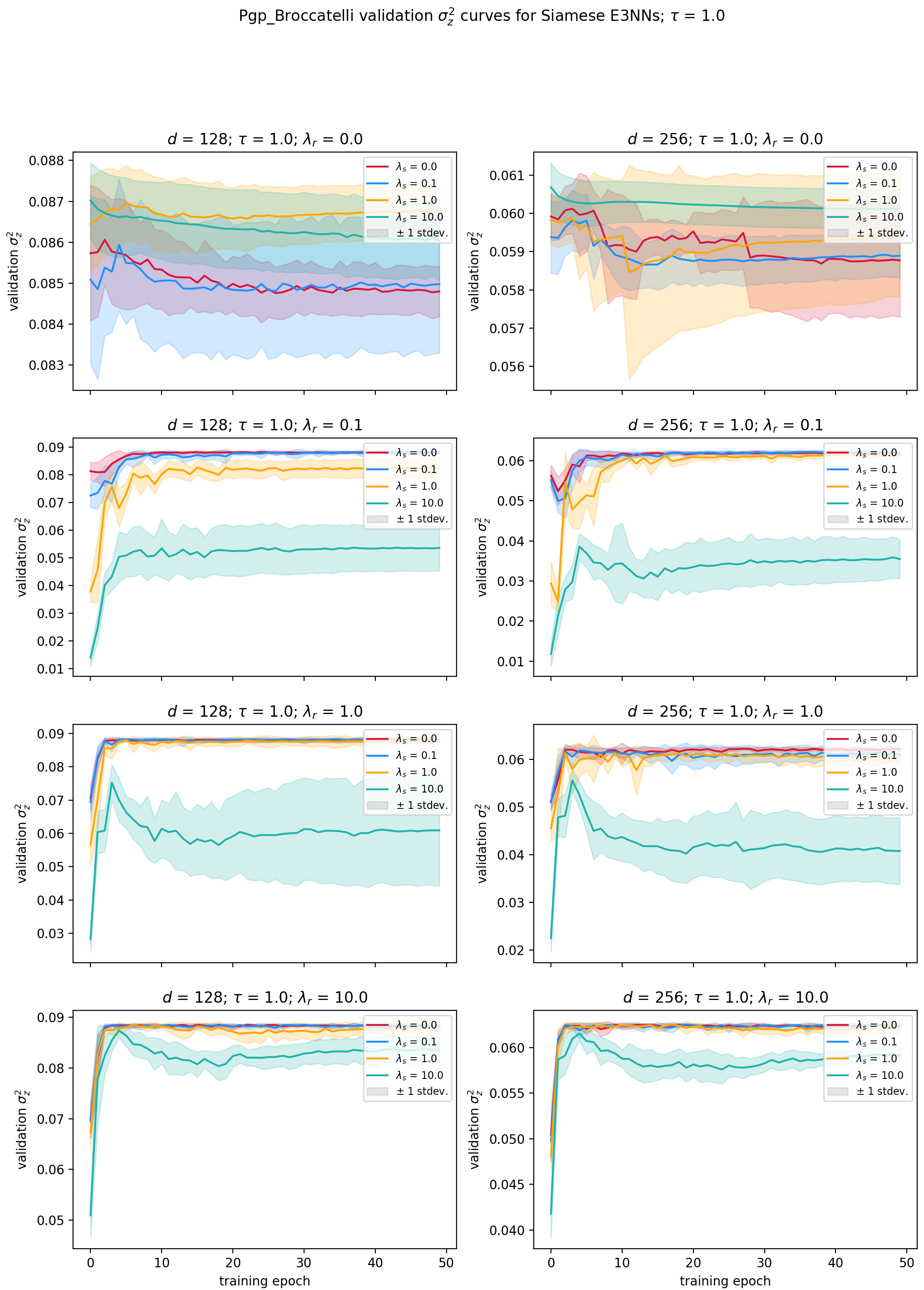}
    %\caption{}
    \label{app:fig:pgp_sig_1.0}
\end{figure}

\begin{figure}[t]
    \centering
    \includegraphics[width=0.98\textwidth]{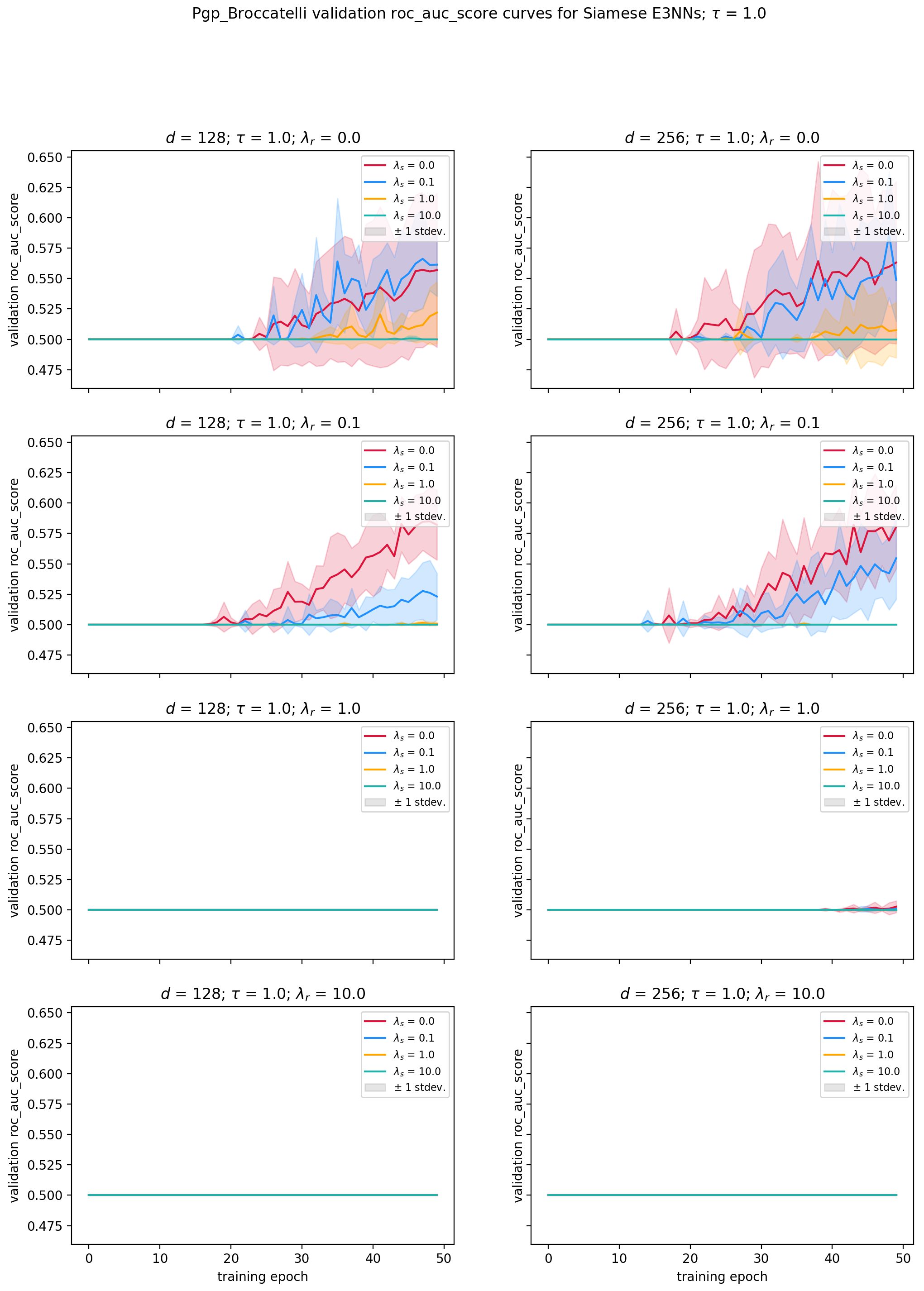}
    %\caption{}
    \label{app:fig:pgp_roc_1.0}
\end{figure}

\begin{figure}[t]
    \centering
    \includegraphics[width=0.98\textwidth]{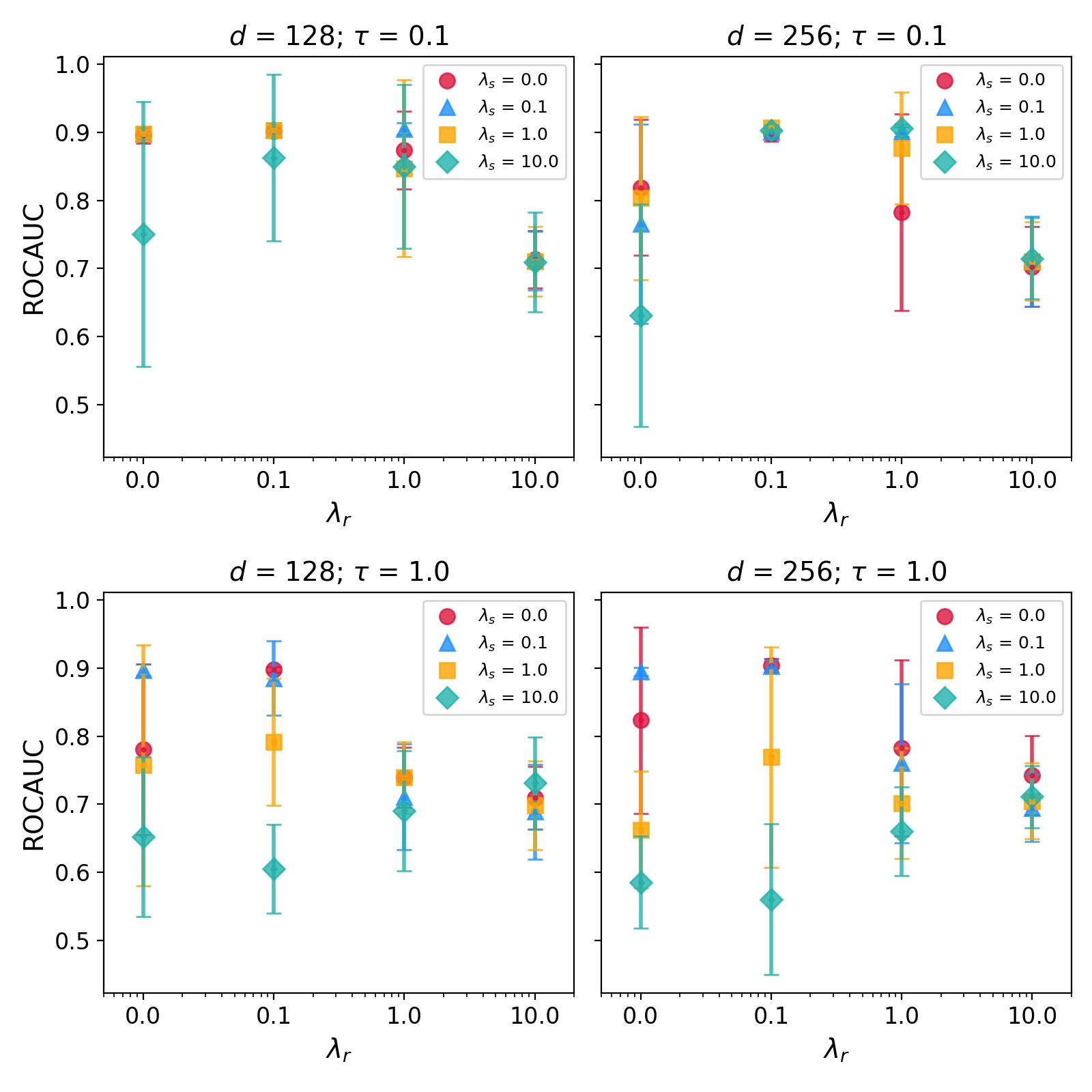}
    %\caption{}
    \label{app:fig:pgp_scatter}
\end{figure}

\clearpage

\subsubsection{\textbf{Clear}}

\begin{figure}[H]
    \centering
    \includegraphics[width=0.98\textwidth]{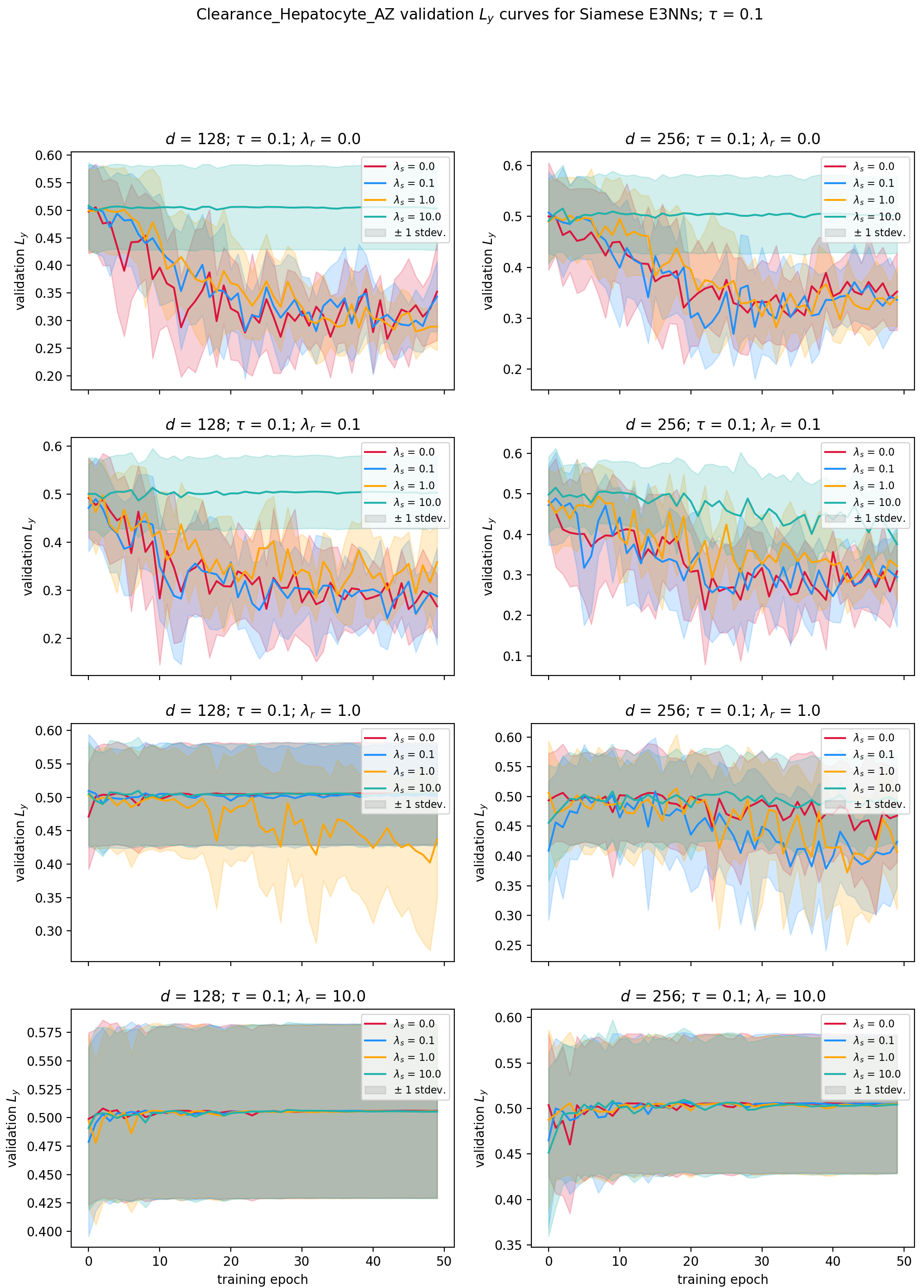}
    %\caption{}
    \label{app:fig:clear_ly_0.1}
\end{figure}

\begin{figure}[t]
    \centering
    \includegraphics[width=0.98\textwidth]{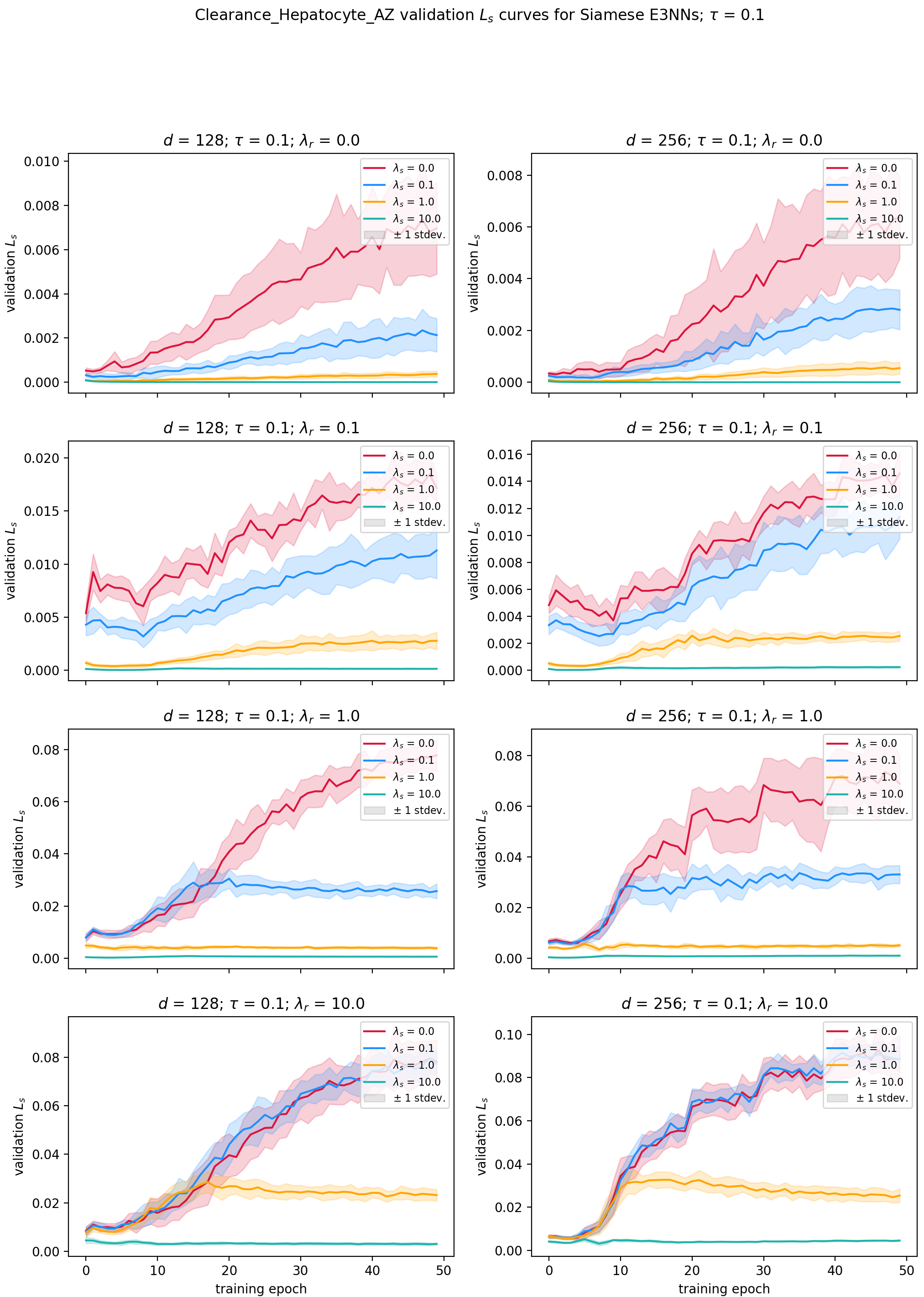}
    %\caption{}
    \label{app:fig:clear_ls_0.1}
\end{figure}

\begin{figure}[t]
    \centering
    \includegraphics[width=0.98\textwidth]{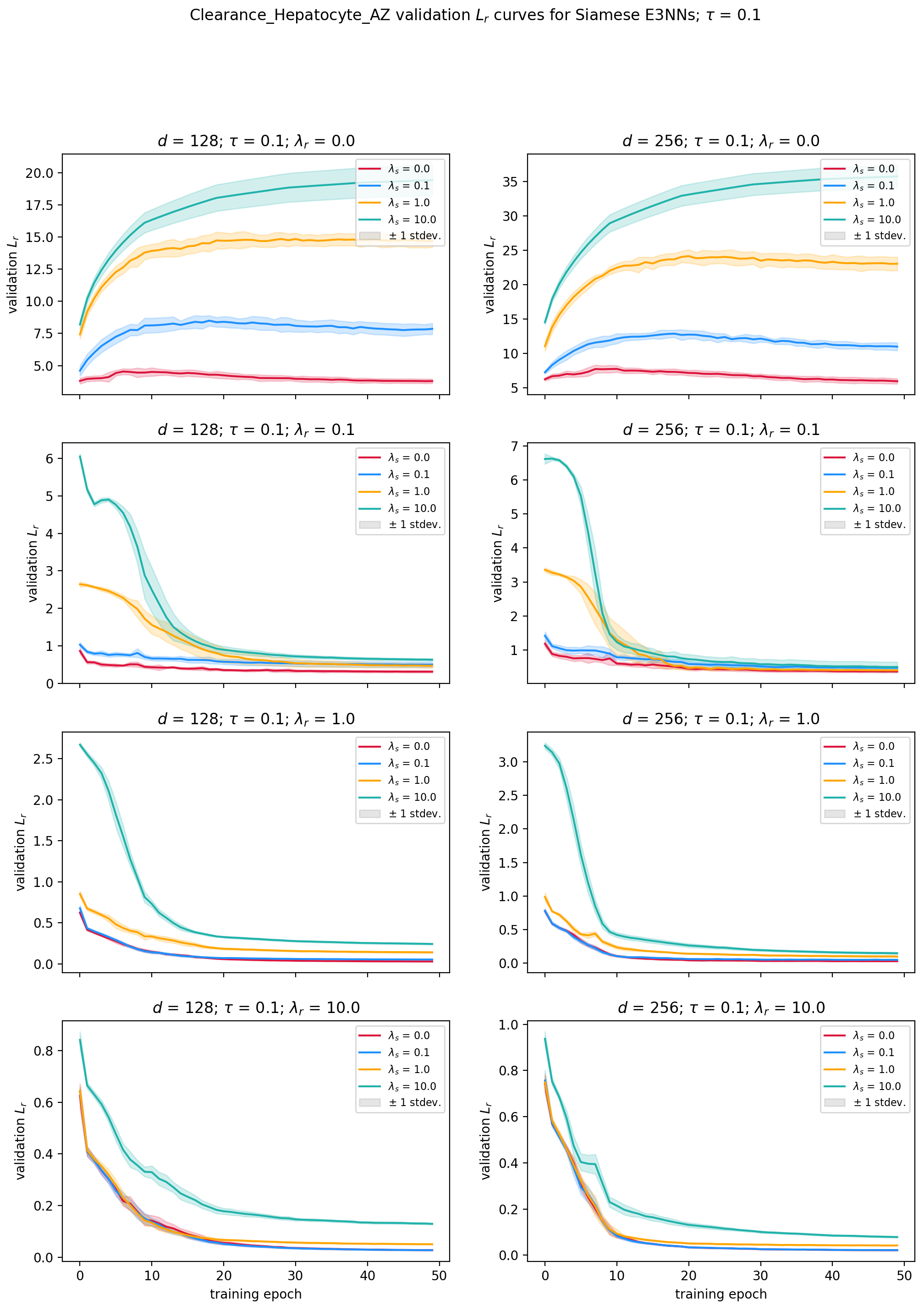}
    %\caption{}
    \label{app:fig:clear_lr_0.1}
\end{figure}

\begin{figure}[t]
    \centering
    \includegraphics[width=0.98\textwidth]{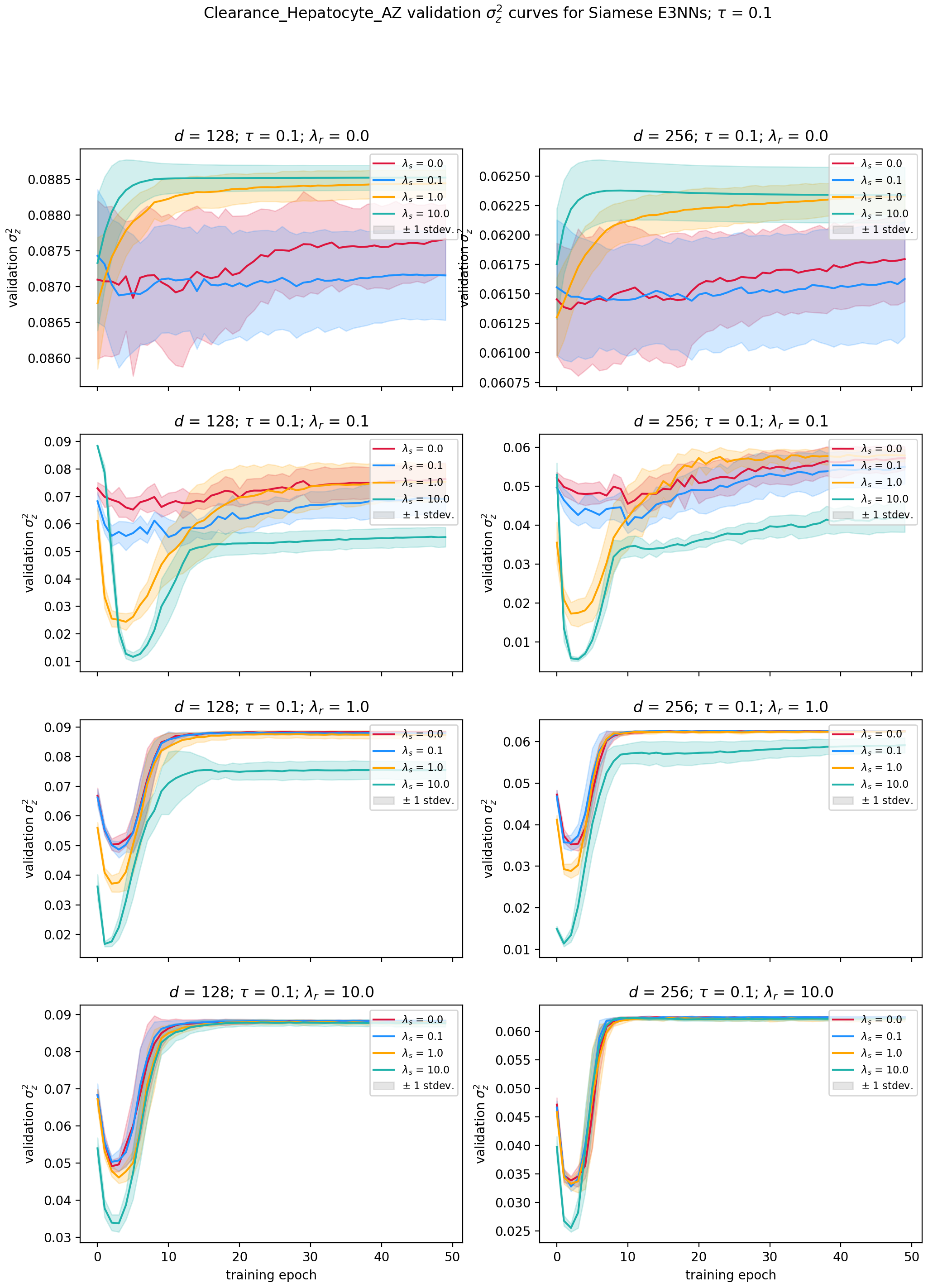}
    %\caption{}
    \label{app:fig:clear_sig_0.1}
\end{figure}

\begin{figure}[t]
    \centering
    \includegraphics[width=0.98\textwidth]{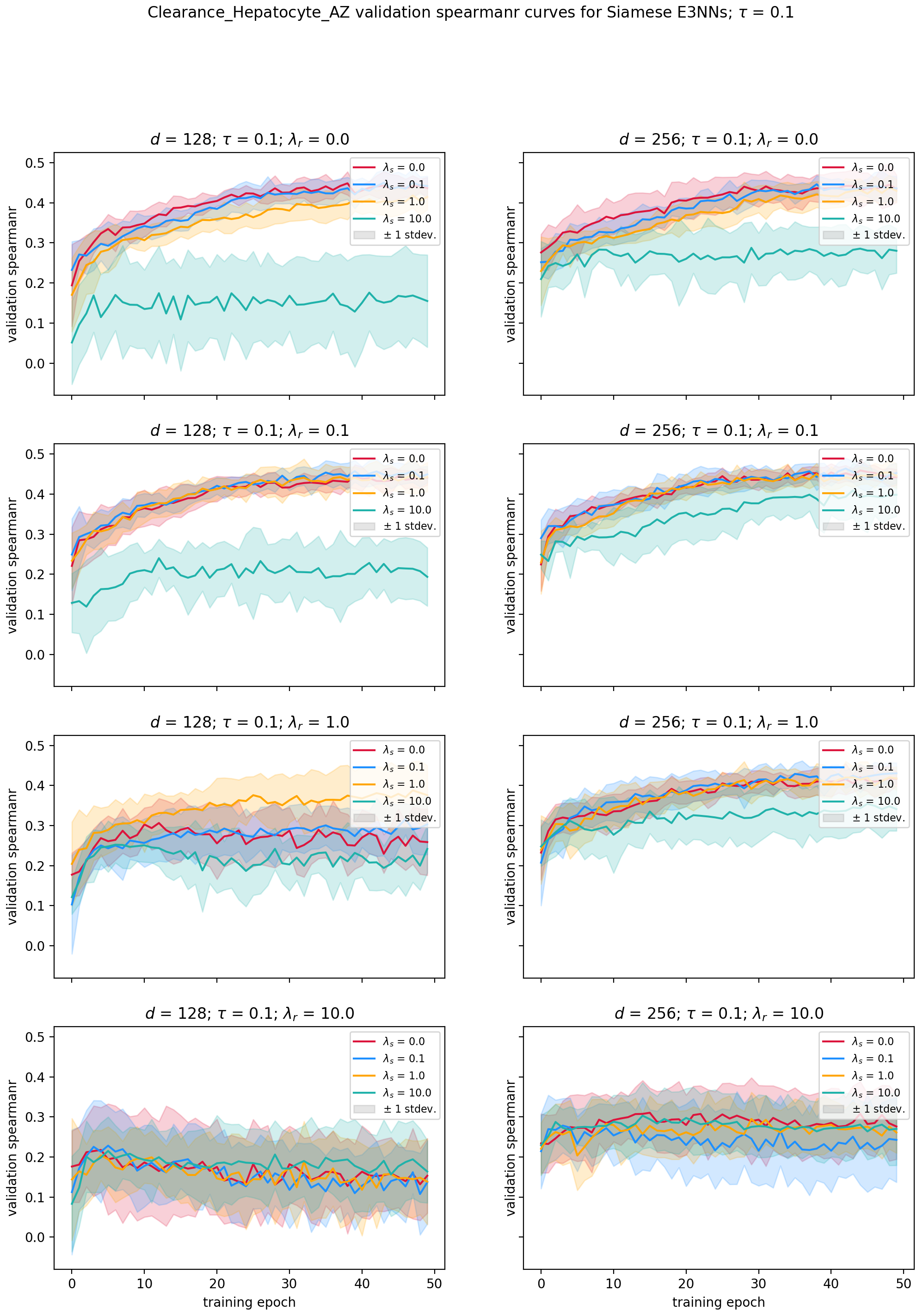}
    %\caption{}
    \label{app:fig:clear_rho_0.1}
\end{figure}

\begin{figure}[t]
    \centering
    \includegraphics[width=0.98\textwidth]{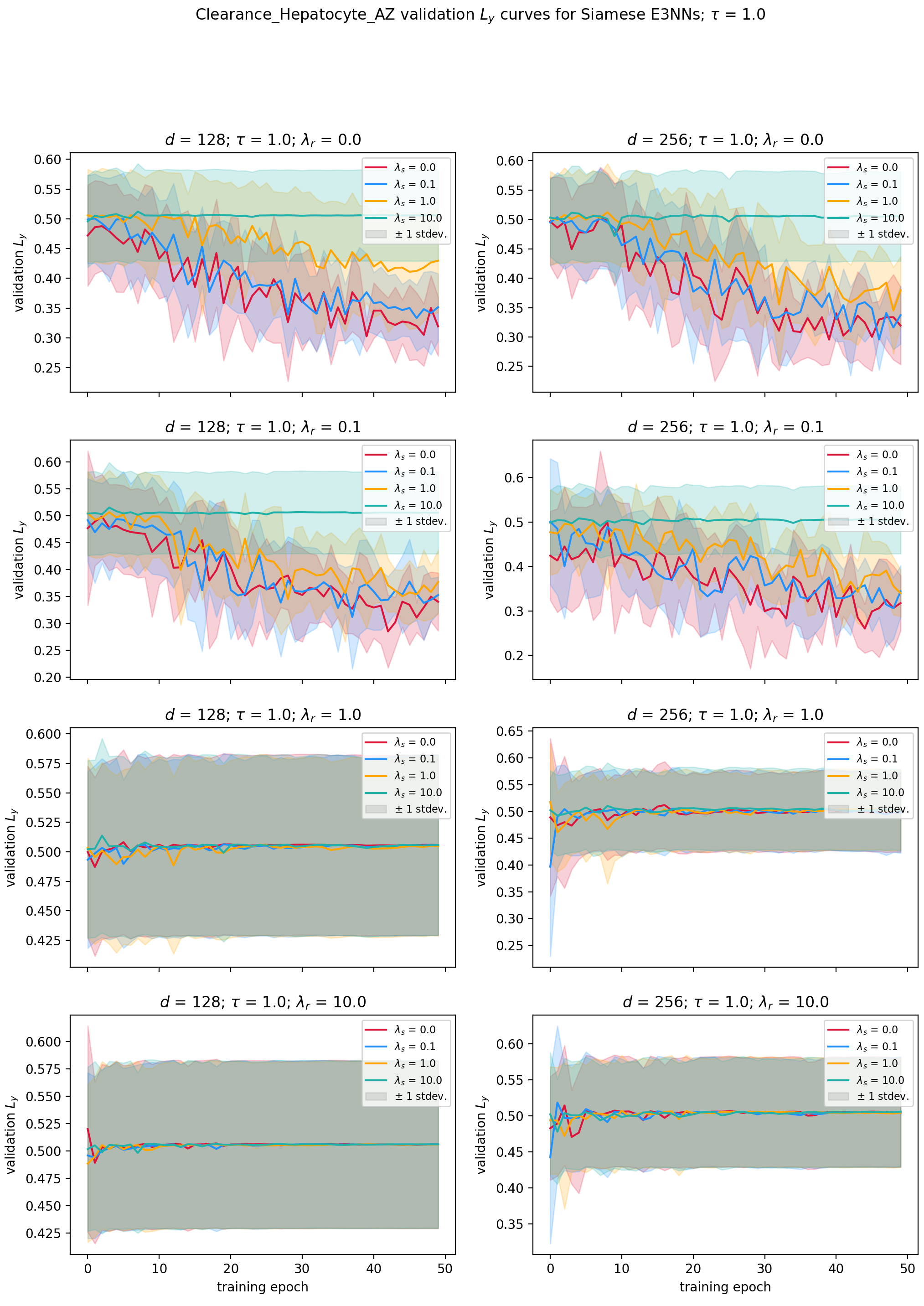}
    %\caption{}
    \label{app:fig:clear_ly_1.0}
\end{figure}

\begin{figure}[t]
    \centering
    \includegraphics[width=0.98\textwidth]{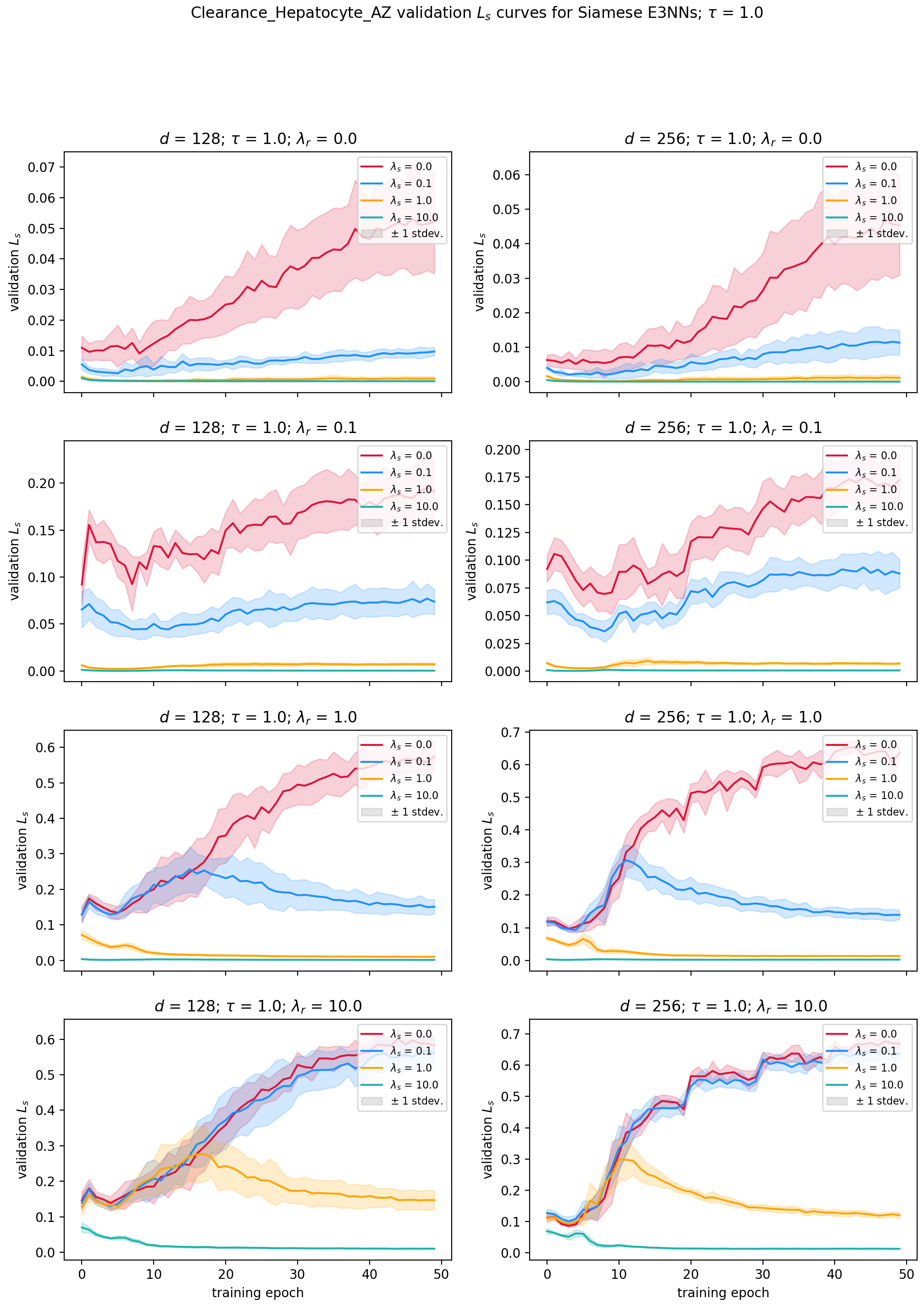}
    %\caption{}
    \label{app:fig:clear_ls_1.0}
\end{figure}

\begin{figure}[t]
    \centering
    \includegraphics[width=0.98\textwidth]{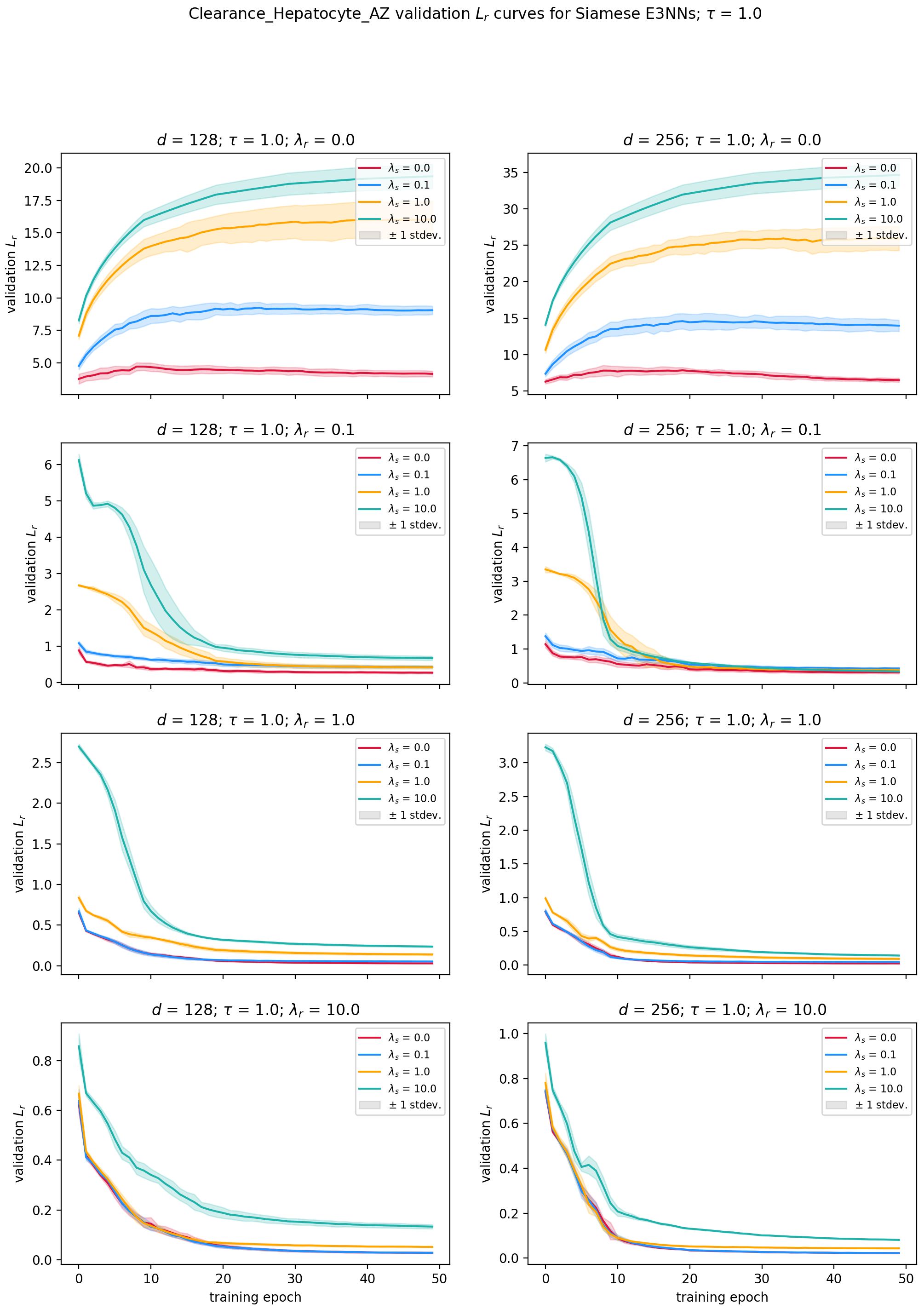}
    %\caption{}
    \label{app:fig:clear_lr_1.0}
\end{figure}

\begin{figure}[t]
    \centering
    \includegraphics[width=0.98\textwidth]{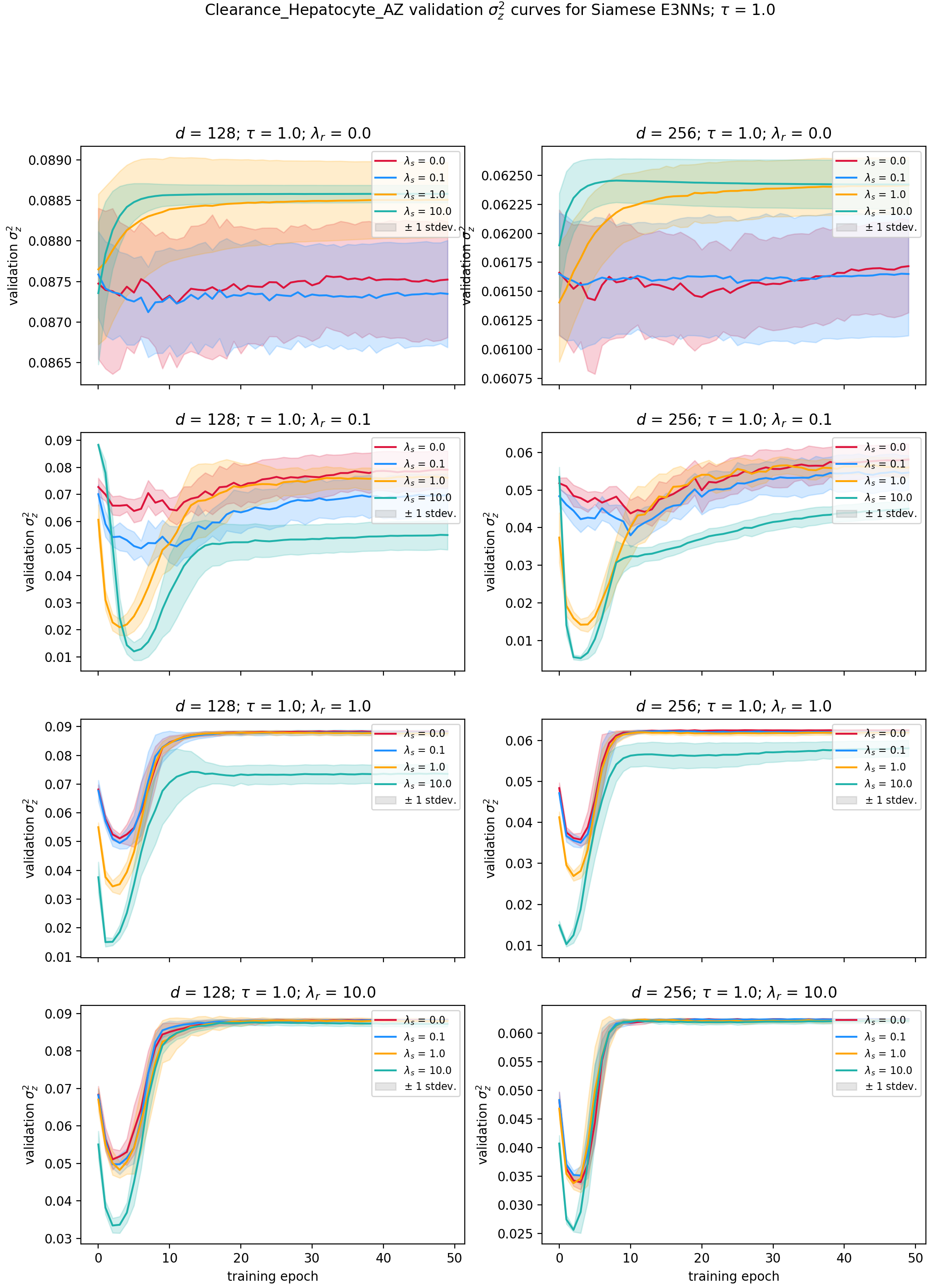}
    %\caption{}
    \label{app:fig:clear_sig_1.0}
\end{figure}

\begin{figure}[t]
    \centering
    \includegraphics[width=0.98\textwidth]{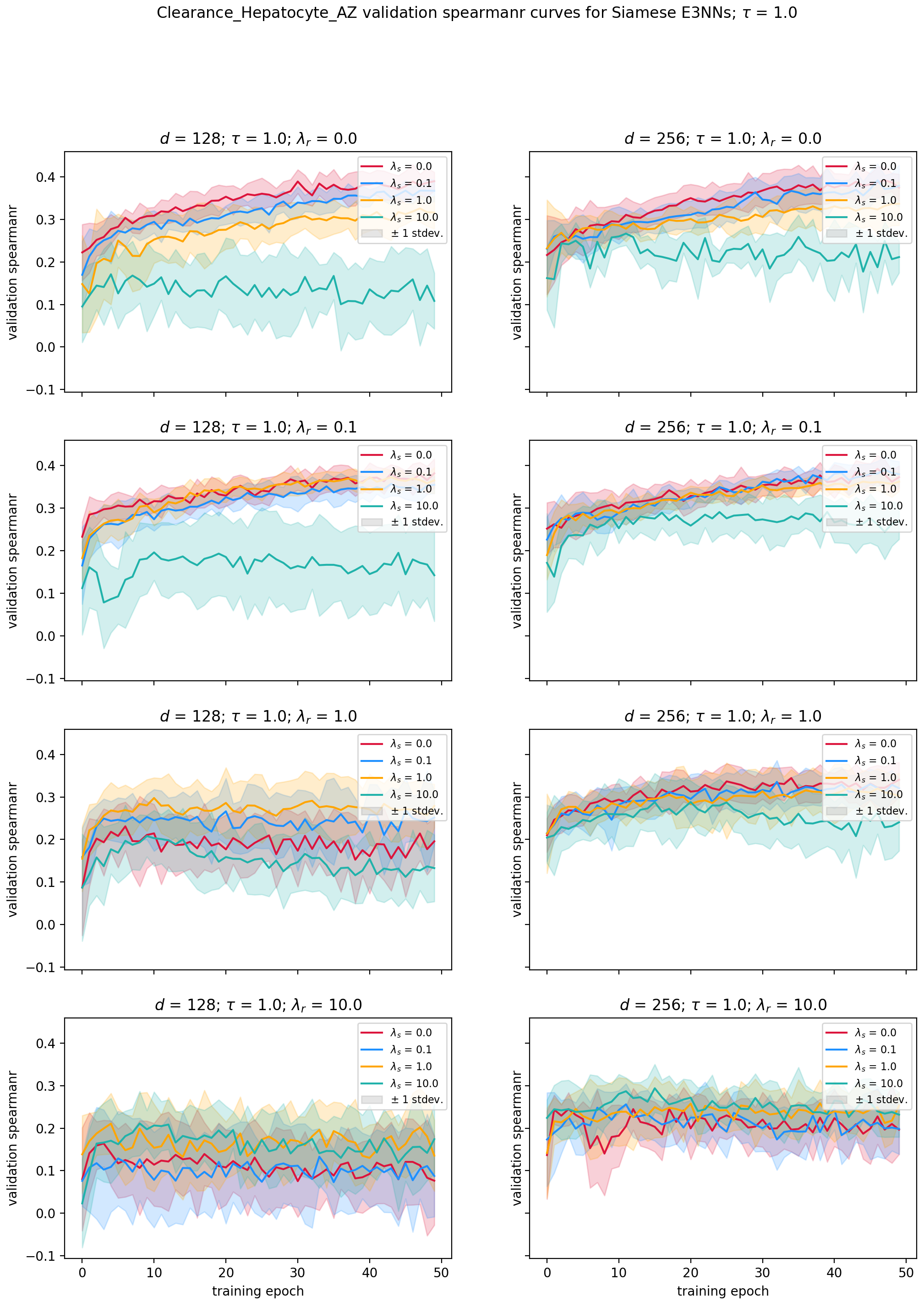}
    %\caption{}
    \label{app:fig:clear_rho_1.0}
\end{figure}

\begin{figure}[t]
    \centering
    \includegraphics[width=0.98\textwidth]{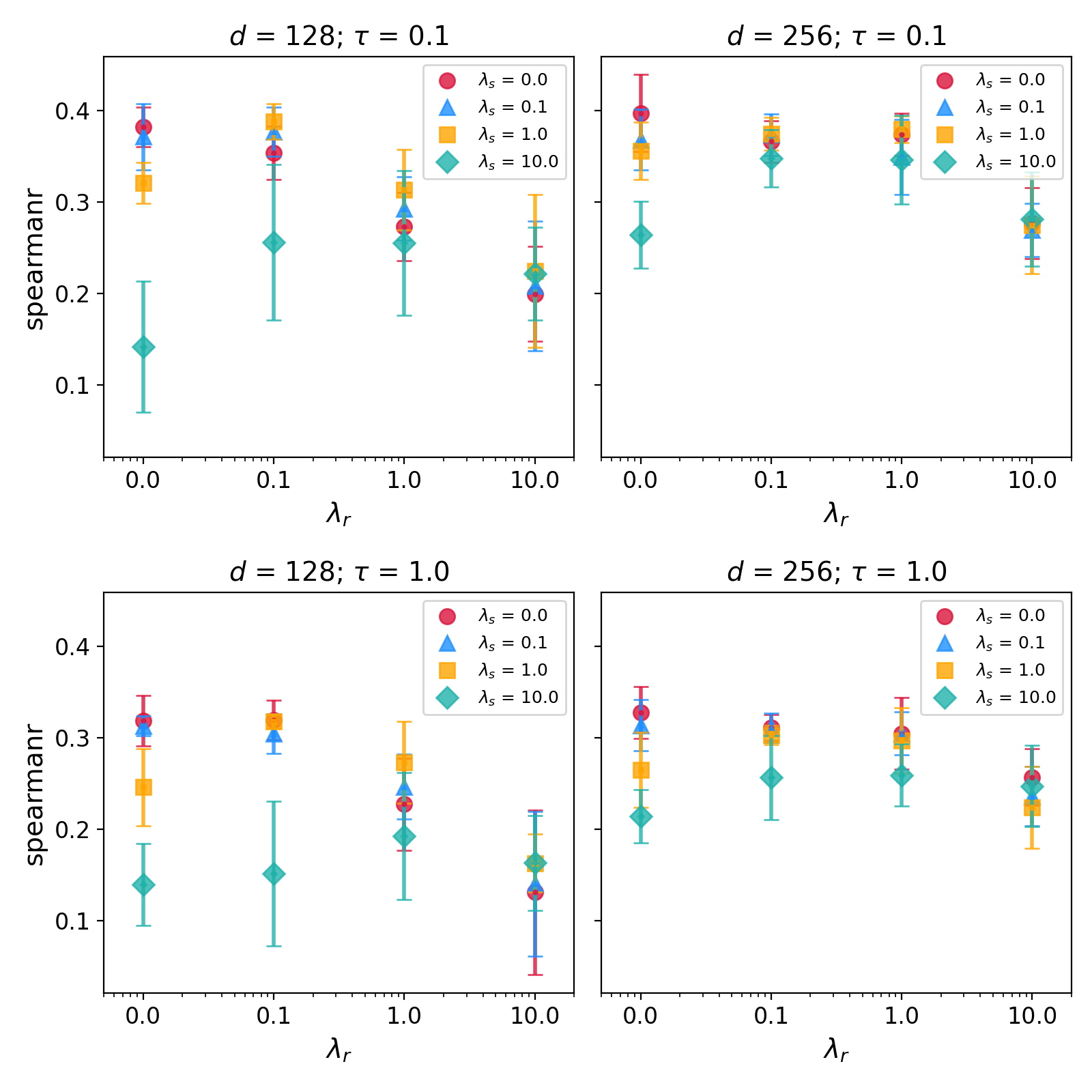}
    %\caption{}
    \label{app:fig:clear_scatter}
\end{figure}

%\end{document}

\clearpage

\subsubsection{Manifold smoothness and partial dimensional collapse}\label{app:sec:ms_pdc}

See Sections \ref{sec:analysis} and \ref{sec:pdc_ms} for details on the analysis of manifold properties. Expanded results from Figure \ref{fig:ms_pdc} are included below, organized in the same manner as those in Section \ref{app:sec:training}. All results shown were computed for test set samples only, which were never seen in training or validation.

\subsubsection{\textbf{Pgp}}

\begin{figure}[H]
    \centering
    \includegraphics[width=0.92\textwidth]{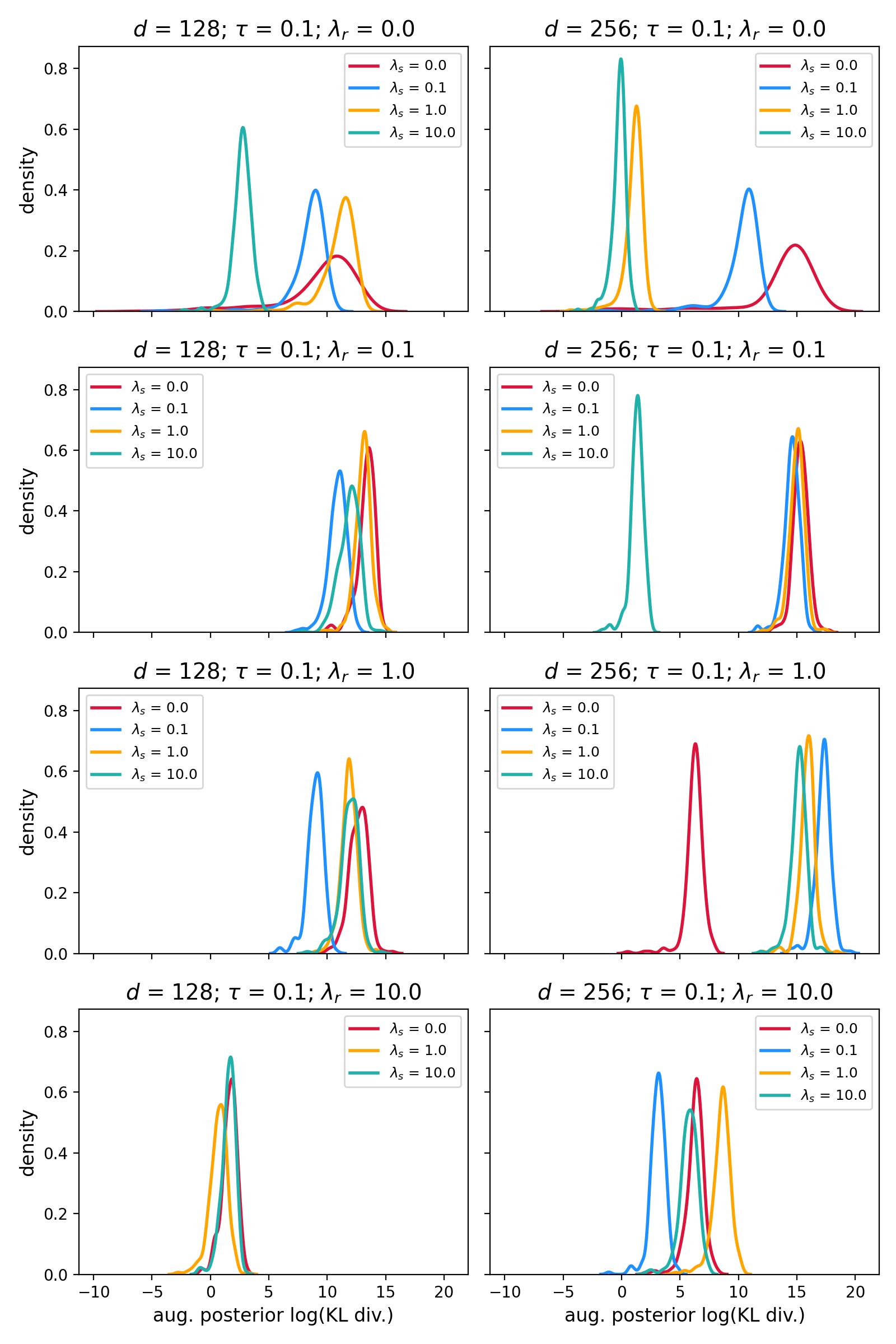}
    %\caption{}
    \label{app:fig:pgp_ms_0.1}
\end{figure}

\begin{figure}[t]
    \centering
    \includegraphics[width=0.92\textwidth]{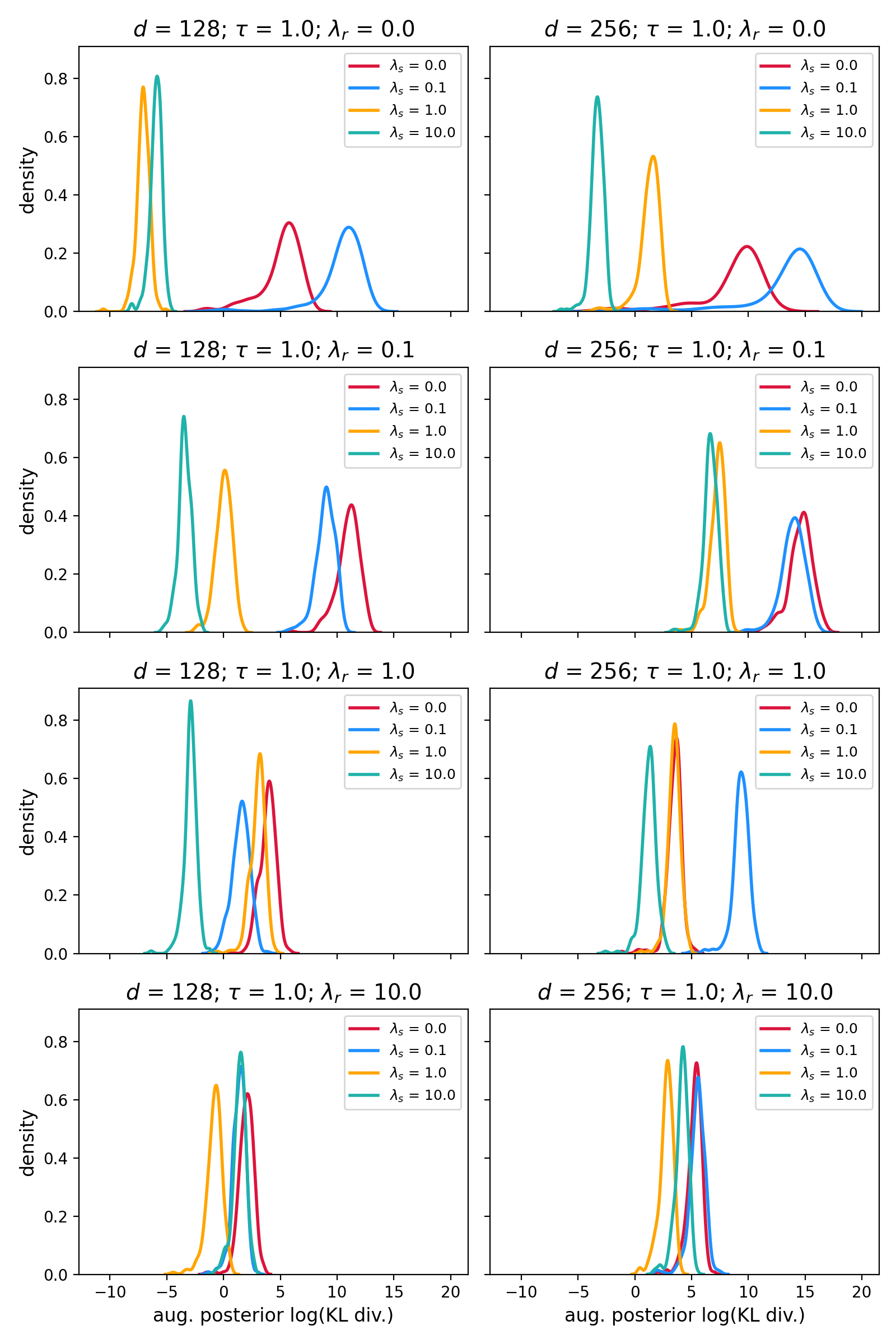}
    %\caption{}
    \label{app:fig:pgp_ms_1.0}
\end{figure}

\begin{figure}[t]
    \centering
    \includegraphics[width=0.92\textwidth]{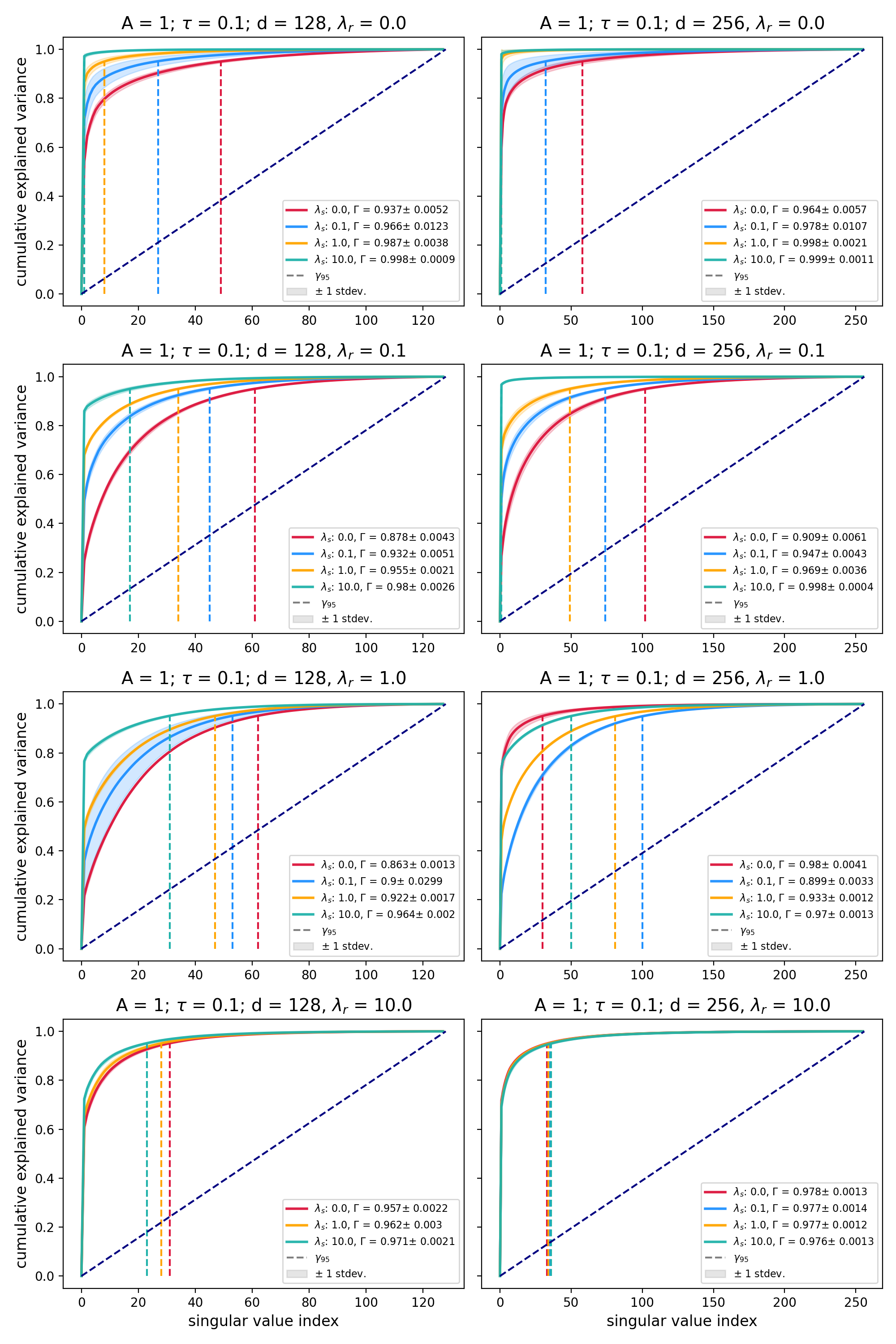}
    %\caption{}
    \label{app:fig:pgp_pdc_0.1}
\end{figure}

\begin{figure}[t]
    \centering
    \includegraphics[width=0.92\textwidth]{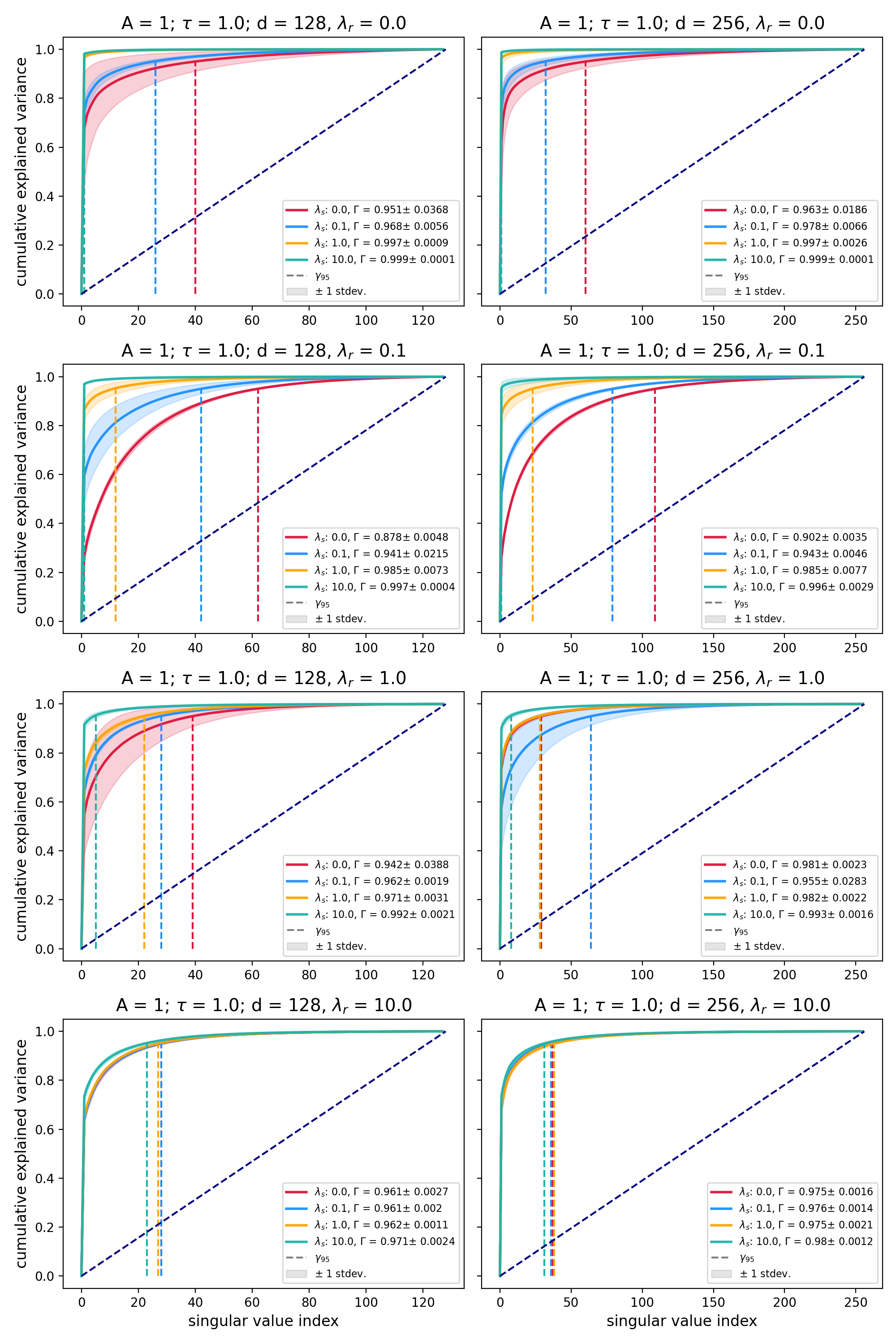}
    %\caption{}
    \label{app:fig:pgp_pdc_1.0}
\end{figure}

\clearpage

\subsubsection{\textbf{Clear}}

\begin{figure}[H]
    \centering
    \includegraphics[width=0.92\textwidth]{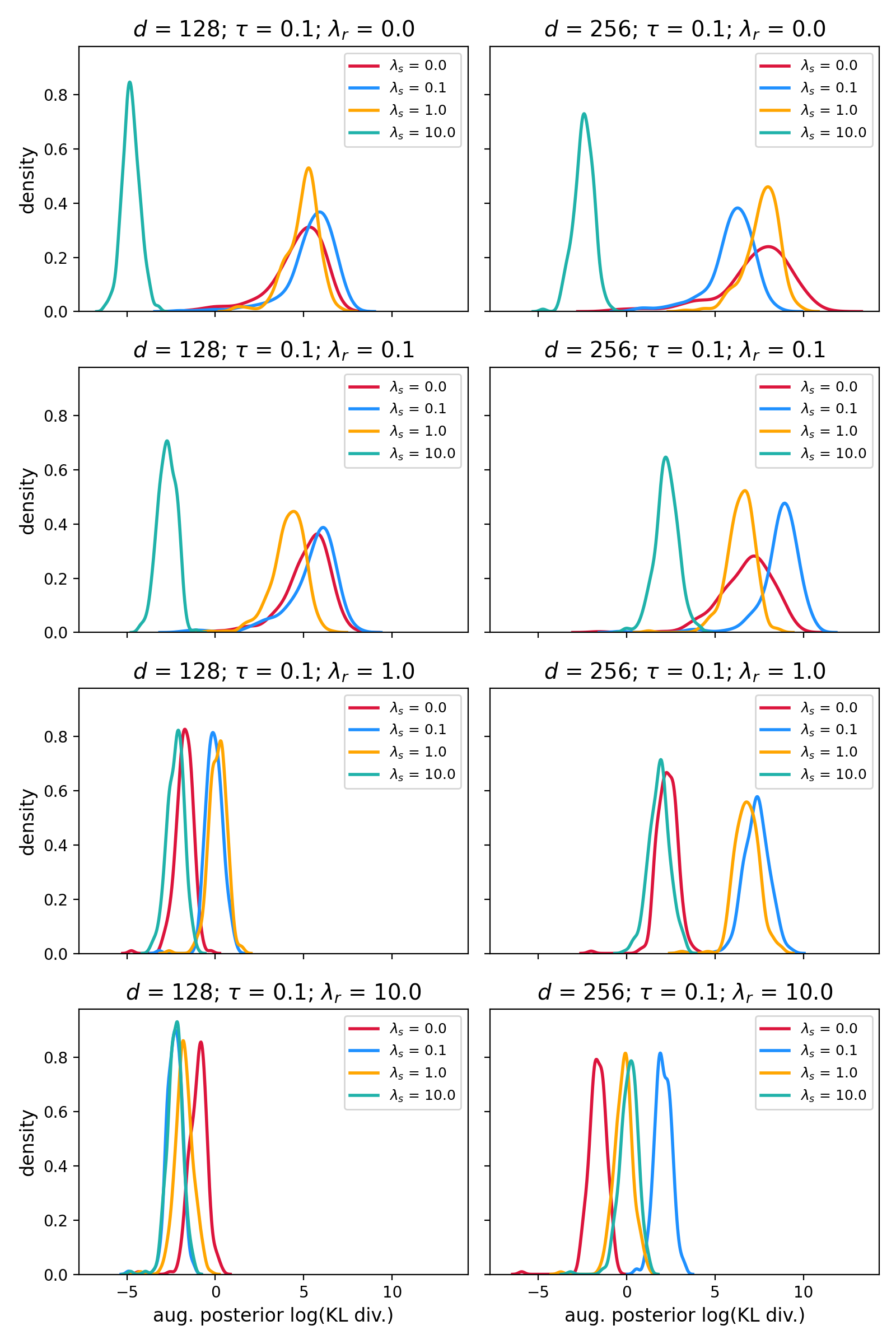}
    %\caption{}
    \label{app:fig:clear_ms_0.1}
\end{figure}

\begin{figure}[t]
    \centering
    \includegraphics[width=0.92\textwidth]{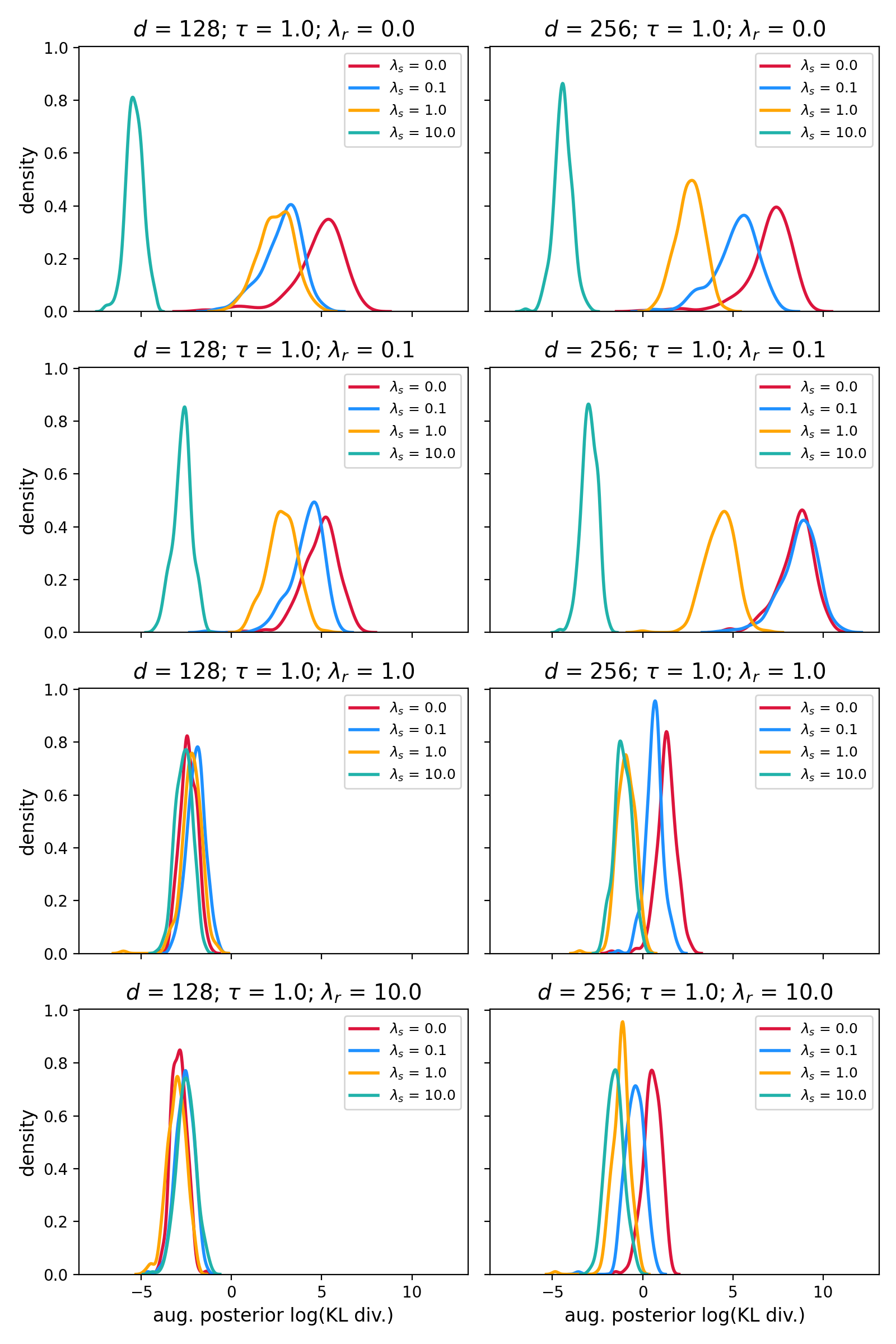}
    %\caption{}
    \label{app:fig:clear_ms_1.0}
\end{figure}

\begin{figure}[t]
    \centering
    \includegraphics[width=0.92\textwidth]{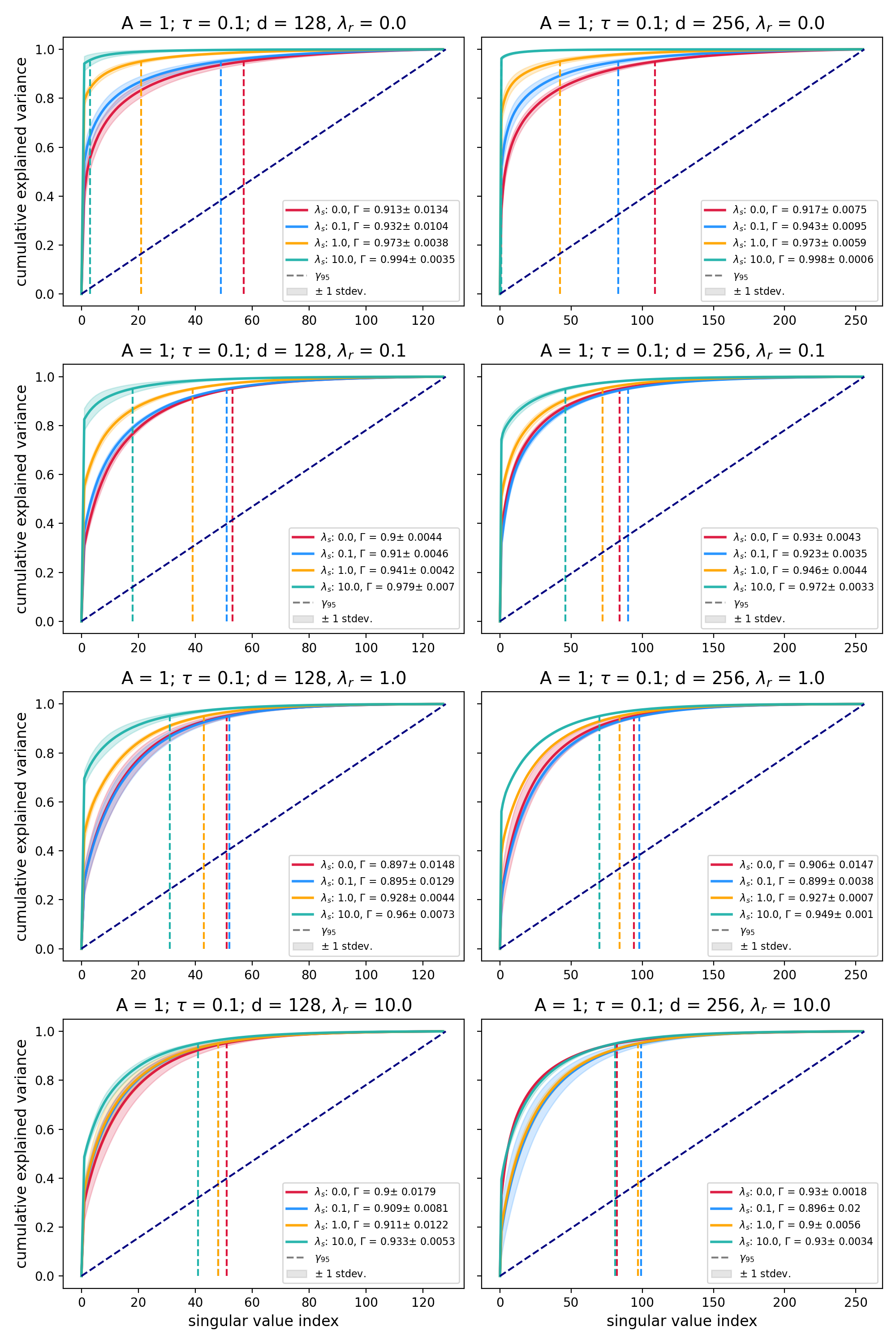}
    %\caption{}
    \label{app:fig:clear_pdc_0.1}
\end{figure}

\begin{figure}[t]
    \centering
    \includegraphics[width=0.92\textwidth]{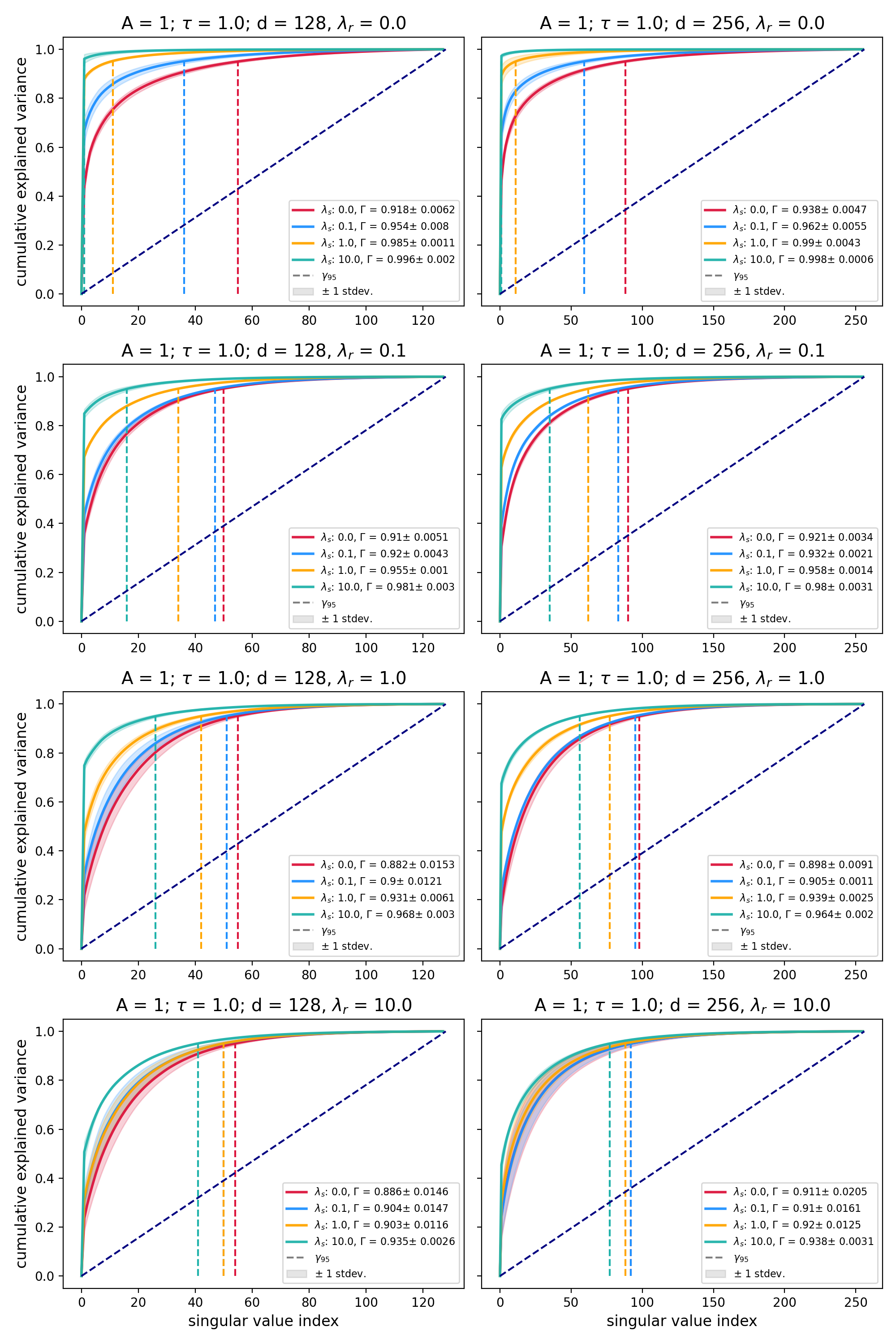}
    %\caption{}
    \label{app:fig:clear_pdc_1.0}
\end{figure}

\end{document}